\documentclass{article}

\PassOptionsToPackage{numbers, compress}{natbib}

    \usepackage[final]{neurips_2024}

\usepackage[utf8]{inputenc} 
\usepackage[T1]{fontenc}    
\usepackage{url}            
\usepackage{booktabs}       
\usepackage{amsfonts}       
\usepackage{nicefrac}       
\usepackage{microtype}      
\usepackage{xcolor}         

\usepackage{amsmath}
\usepackage{graphicx} 
\usepackage{subcaption}
\usepackage{multirow}
\usepackage{wrapfig}
\usepackage{subcaption}
\usepackage[unicode=true,
          colorlinks=true]{hyperref}

\title{GSGAN: Adversarial Learning for \\ Hierarchical Generation of 3D Gaussian Splats}

\author{
    Sangeek Hyun \\
    Sungkyunkwan University
    \And
    Jae-Pil Heo\thanks{Corresponding author} \\
    Sungkyunkwan University
}

\begin{document}

\maketitle

\begin{abstract}
Most advances in 3D Generative Adversarial Networks (3D GANs) largely depend on ray casting-based volume rendering, which incurs demanding rendering costs. One promising alternative is rasterization-based 3D Gaussian Splatting (3D-GS), providing a much faster rendering speed and explicit 3D representation. In this paper, we exploit Gaussian as a 3D representation for 3D GANs by leveraging its efficient and explicit characteristics. However, in an adversarial framework, we observe that a na\"ive generator architecture suffers from training instability and lacks the capability to adjust the scale of Gaussians. This leads to model divergence and visual artifacts due to the absence of proper guidance for initialized positions of Gaussians and densification to manage their scales adaptively. 
To address these issues, we introduce GSGAN, a generator architecture with a hierarchical multi-scale Gaussian representation that effectively regularizes the position and scale of generated Gaussians. 
Specifically, we design a hierarchy of Gaussians where finer-level Gaussians are parameterized by their coarser-level counterparts; the position of finer-level Gaussians would be located near their coarser-level counterparts, and the scale would monotonically decrease as the level becomes finer, modeling both coarse and fine details of the 3D scene. Experimental results demonstrate that ours achieves a significantly faster rendering speed (×100) compared to state-of-the-art 3D consistent GANs with comparable 3D generation capability.
Project page: \href{https://hse1032.github.io/gsgan}{https://hse1032.github.io/gsgan}.
\end{abstract}

\vspace{-0.5cm}
\section{Introduction}
\label{sec:1_intro}

\begin{figure}[h]
    \vspace{-0.1cm}
    \centering
    \includegraphics[width=1.0\columnwidth]{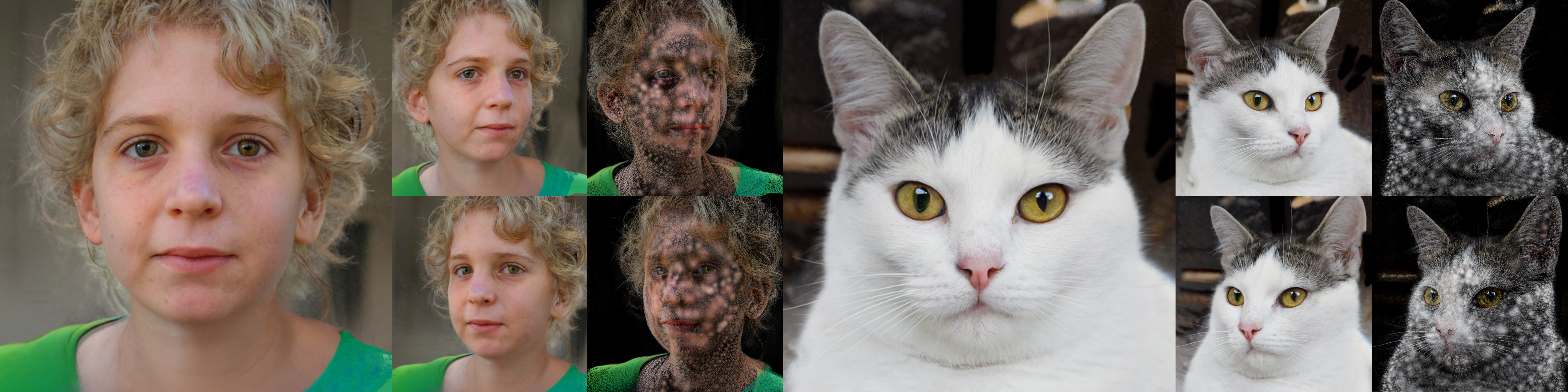}
    \caption{
    Generated examples from the proposed method~(FFHQ-512, AFHQ-Cat-512). 
    Ours synthesize multi-view consistent images with a significantly faster rendering speed by leveraging 3D Gaussian representation. 
    We represent a 3D scene as a composite of hierarchical Gaussians, where each level of Gaussian depicts coarse and fine details corresponding to its level.
    To visualize the effects of individual Gaussian, the right-most images are rendered by reducing the scale of Gaussians. 
    }
    \label{fig: intro example}
    \vspace{-0.1cm}
\end{figure}

The research field of 3D generative models has recently emerged and shows impressive generation capability in various domains such as text-to-3D~\cite{jain2022zero,poole2022dreamfusion,tang2023dreamgaussian,wang2023prolificdreamer,lin2023magic3d,wang2023score,chen2023fantasia3d} and image-to-3D~\cite{raj2023dreambooth3d,tang2023make,melas2023realfusion}. Among them, 3D Generative Adversarial Networks~\cite{chan2021pi,chan2022efficient,deng2022gram,gu2021stylenerf,skorokhodov2022epigraf,nguyen2019hologan} are an adversarial learning framework between a 3D generator and a 2D discriminator, capable of synthesizing 3D models solely by training with collections of 2D images or with the additional use of their corresponding camera poses. Specifically, the generator synthesizes a 3D model and renders it as a 2D image using given camera parameters, and then the 2D discriminator determines its realness and its match to the given camera pose.

The generally used volume rendering in 3D GANs is ray casting~\cite{kajiya1984ray}, which is widely adopted in NeRF literature~\cite{mildenhall2021nerf,barron2021mip,barron2022mip,muller2022instant}. Ray casting-based rendering demonstrates its prominent capability to model a 3D scene in various domains of research; however, it is also known for its excessive computational costs. Specifically, it requires $(H \times W \times D)$ point sampling for rendering an image from a single camera view, where $H, W, D$ denotes height, width, the number of sampled points on a ray~(depth).
This demanding computational cost hinders previous methods in 3D GANs from performing the rendering process at higher resolutions~\cite{chan2022efficient} or forces them to use architectures optimized for efficient 3D representation, leading to degraded generation quality~\cite{deng2022gram,skorokhodov2022epigraf}.

Recently, 3D Gaussian Splatting (3D-GS)~\cite{kerbl20233d} has been introduced for the 3D reconstruction task. This method represents a 3D scene as a composition of 3D Gaussians, projecting and $\alpha$-blending these Gaussians to render images. This rasterization-based method notably enhances rendering speed. Despite this innovative progress in the 3D reconstruction task, its application in 3D generative models has not yet been studied. One important challenge is the training algorithm of 3D-GS. Unlike the scene-independent and fully differentiable training of NeRFs, 3D-GS requires additional constraints such as the proper initialization of Gaussian positions for a given scene by SfM~\cite{snavely2006photo} and densification that heuristically manages the scale and number of Gaussians. Since 3D GANs deploy a single generator trained by gradient descent to model the distribution of 3D data, these non-differentiable and scene-dependent characteristics of 3D-GS restrict its application in 3D GANs.

In this paper, we extend the application of 3D Gaussian representation with rasterization to 3D GANs, leveraging its efficient rendering speed for high-resolution data. 
In an adversarial learning framework, we observe that a na\"ive generator architecture, which simply synthesizes a set of Gaussians without any constraints, suffers from training instability and imprecise adjustment of the scale of Gaussians. 
For example, at the early stage of training, all the Gaussians disappear in the rendered images, or there are visual artifacts created by long-shaped Gaussians despite its convergence.

To address these training difficulties, we devise a method to regularize the Gaussian representation for 3D GANs, focusing particularly on the position and scale parameters. To this end, we propose a generator architecture with a hierarchical Gaussian representation. This hierarchical representation models the Gaussians of adjacent levels to be dependent, encouraging the generator to synthesize the 3D space in a coarse-to-fine manner.
Specifically, we first introduce a locality constraint whereby the positions of fine-level Gaussians are located near their coarse-level counterpart Gaussians and are parameterized by them, thus reducing the possible positions of newly added Gaussians. Then, we design the scale of Gaussians to monotonically decrease as the level of Gaussians becomes finer, facilitating the generator's ability to model the scene in both coarse and fine details.
Based on this representation, we propose a generator architecture that effectively implements the hierarchy of Gaussians. 
With additional architectural details, we validate that the proposed generator successfully synthesizes realistic 3D models~(Fig.~\ref{fig: intro example}) with stabilized training and enhanced generation quality.

Our contributions are summarized as follows:
\begin{itemize}
\item[-] We firstly exploit a rasterization-based 3D Gaussian representation for efficient 3D GANs.
\item[-] We introduce GSGAN, a hierarchical 3D Gaussian representation that regularizes the parameters of Gaussians by building dependencies between Gaussians of adjacent hierarchies, stabilizing the training of Gaussian splatting-based 3D GANs.
\item[-] The proposed method achieves a significantly faster rendering speed compared to state-of-the-art 3D GANs while maintaining comparable generation quality and 3D consistency.
\end{itemize}

\section{Related works}
\label{sec:2_related_work}

\subsection{3D Generative Adversarial Networks}
3D GANs~\cite{chan2021pi,nguyen2019hologan} learn the 3D representation from a collection of 2D images with corresponding camera parameters. Recent progress mostly focuses on generator architecture, especially for enhancing efficiency. For example, EG3D~\cite{chan2022efficient} introduces the tri-plane representation with an additional upsampling 2D image after volume rendering. Differently, there is another stream that directly renders a high-resolution image without a 2D upsampler. For instance, GRAM~\cite{deng2022gram} learns the iso-surfaces for efficient sampling of points on a ray, based on an implicit-network generator~\cite{chan2021pi}. 
Additionally, Voxgraf~\cite{schwarz2022voxgraf} exploits the sparse voxel grid instead of feature fields to represent the 3D scene. Epigraf~\cite{skorokhodov2022epigraf} introduces a patch-wise training scheme to reduce the computational cost of volume rendering at high resolution (e.g., 512$\times$512) during the training phase. 
Differently, Mimic3D~\cite{chen2023mimic3d} proposes a method to learn 3D representation at high resolution by distilling the 2D upsampling result of EG3D into the high-resolution tri-plane.
Most recently, WYSIWIG~\cite{trevithick2024you} proposes SDF representation-based GANs for reducing the number of sampling points on rays in a high-resolution 3D representation.

While previous methods use ray casting as a volume rendering method with demanding rendering costs, we exploit efficient rasterization by adopting 3D Gaussians as a 3D representation.

\subsection{Gaussian splatting for 3D representation}
3D Gaussian Splatting (3D-GS)~\cite{kerbl20233d} represents a 3D scene as a composition of 3D Gaussians. This approach is known for its fast rendering speed, which rasterization-based rendering leads to, as well as its prominent reconstruction quality and faster convergence. 
Recently, various domains of research utilize 3D-GS, such as human avatar modeling\cite{qian2023gaussianavatars,li2023animatable,xu2023gaussian}, facial editing~\cite{zhou2024headstudio}, text-to-3D synthesis~\cite{jain2022zero,poole2022dreamfusion,tang2023dreamgaussian,wang2023prolificdreamer,lin2023magic3d,wang2023score,chen2023fantasia3d}, and image-to-3D synthesis~\cite{raj2023dreambooth3d,tang2023make,melas2023realfusion}. 
These approaches typically substitute the neural radiance field with a Gaussian representation to leverage its efficient characteristics.

The research efforts also try to apply 3D Gaussian representation to an adversarial learning framework, especially with structural prior such as the human body and facial template.
For example, Gaussian Shell Maps~\cite{abdal2024gaussian} focus on the task of 3D human generation based on SMPL template~\cite{loper2023smpl} with an adversarial learning framework. Given a human body template, the generator learns to synthesize multiple shell maps containing Gaussian parameters, achieving 3D consistent generation with faster rendering speed.
Similarly, GGHead~\cite{kirschstein2024gghead} focuses on 3D face generation using FLAME template~\cite{FLAME:SiggraphAsia2017}. Similar to Gaussian Shell Maps, they synthesize texture maps containing Gaussian parameters and position offsets by 2D generator based on StyleGAN2~\cite{karras2019style}.
However, these methods largely depend on the pre-defined template as structural prior.

Differently, in this paper, we focus on extending the application of 3D-GS to Generative Adversarial Networks without any geometric prior such as SfM that 3D-GS traditionally requires.

\section{Proposed method}
\label{sec:3_method}

\subsection{Preliminaries}
\label{sec:3.1_backgrounds}
\paragraph{3D Generative Adversarial Networks}
3D GANs~\cite{nguyen2019hologan,chan2021pi,chan2022efficient} is an adversarial learning framework between the 3D generator $g(z, \theta)$ and 2D discriminator $d(I)$.
Specifically, for a given randomly sampled latent code $z \in \mathbb{R}^{d_z} \sim p_z$ and camera pose $\theta \in \mathbb{R}^{d_\theta} \sim p_\theta$, the generator $g(z, \theta)$ synthesizes a 3D scene and renders it to a 2D fake image $I_f \in \mathbb{R}^{H \times W \times 3}$ based on the given camera pose.
Then, the discriminator learns to discriminate the real image $I_r \in \mathbb{R}^{H \times W \times 3}$ and the fake image $I_f$, while the generator learns to deceive the discriminator.
In addition, the discriminator often uses the camera pose to encourage $d$ to be aware of 3D information by making $d$ estimate the camera parameter~\cite{deng2022gram,jo20233d} or performing conditional adversarial loss~\cite{chan2022efficient}.

\paragraph{3D Gaussian splatting}
3D Gaussian representation for 3D modeling is recently introduced by 3D Gaussian Splatting~\cite{kerbl20233d}.
It represents a scene as a composition of anisotropic 3D Gaussians estimated by multi-view images and their corresponding camera poses.
This Gaussian representation contains parameters such as a position $\mu \in \mathbb{R}^{3}$, a scale $s \in \mathbb{R}^{3}$, a quaternion $q \in \mathbb{R}^{4}$, an opacity $\alpha \in \mathbb{R}^{1}$, and a color $c \in \mathbb{R}^{3}$.
In world space, Gaussian is defined as follows:
\begin{equation}
    G(x) = e^{-\frac{1}{2}(x - \mu)^{T} \Sigma^{-1} (x - \mu)}, \quad \Sigma = RSS^TR^T,
\end{equation}
where $R \in \mathbb{R}^{4 \times 4}$ and $S \in \mathbb{R}^{4 \times 4}$ are a rotation matrix and a scaling matrix obtained from a quaternion $q$ and scale $s$, and $x$ is a world coordinate.

To render the image, the color of each pixel $C$ is determined by blending the contributions of all $N$ 3D Gaussians that overlap with the pixel as follows:
\begin{equation}
    C = \sum_{i=1}^{N} c_i \alpha^{\prime}_{i} \prod_{j=1}^{i-1} (1 - \alpha^{\prime}_j), 
\end{equation}
where $c_i$ is the color of each Gaussian, and $\alpha^{\prime}$ is blending weight of 2D projection of Gaussian multiplied by a per-point opacity $\alpha$. The order of Gaussians is sorted by their depth.

\begin{figure}[t!]
     \begin{subfigure}[b]{0.62\textwidth}
        \centering
         \includegraphics[width=\textwidth]{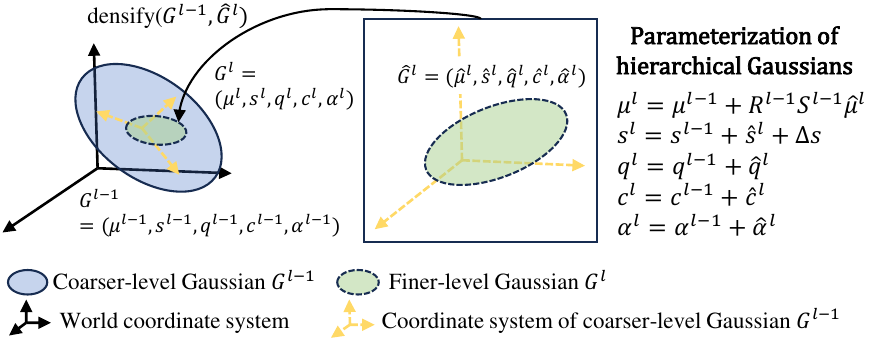}
         \caption{Illustration of hierarchical representation of Gaussians}
    \end{subfigure}
     \begin{subfigure}[b]{0.36\textwidth}
         \centering
         \includegraphics[width=\textwidth]{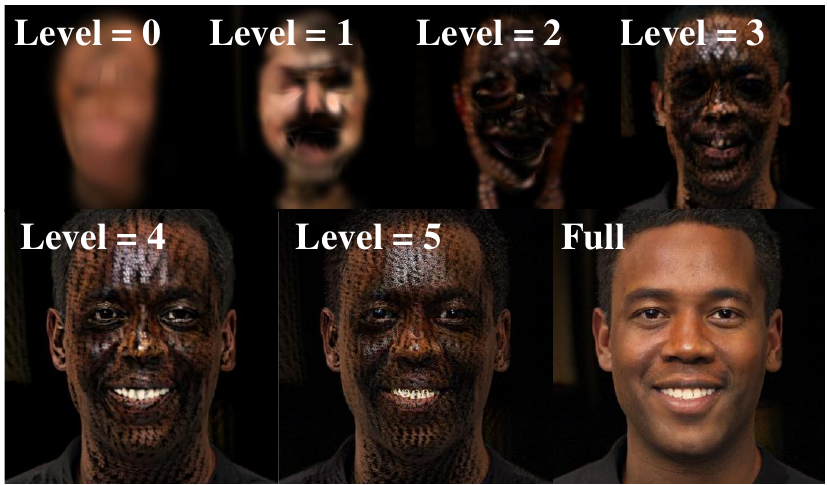} 
         \caption{Example of generated Gaussians of various hierarchy levels}
    \end{subfigure}
    \centering
    \caption{
    Illustration and examples of hierarchical Gaussian representation. 
    (a) We parameterize the finer-level Gaussians by the parameters of coarser-level counterparts for regularizing the scale and position of synthesized Gaussians. 
    (b) Example of synthesized Gaussians across multiple hierarchy levels.
    Gaussians represent coarse or fine details according to its hierarchy level.
    }
    \label{fig: gaussian hierarchy}
\end{figure}

\subsection{Hierarchical 3D Gaussian representation}
\label{sec:hierarchical 3D gaussian sapltting}

Our focus is on utilizing Gaussians as the 3D representation of the generator in 3D GANs. 
We begin with a simple generator that takes a randomly sampled latent code
$z$ as input and outputs $N$ Gaussians, without any restrictions.
However, in this scenario, we observe that this na\"ive application suffers from training instability and fails to properly manage the position $\mu$ and scale $s$, as shown in examples in Fig.~\ref{fig: ablation point clouds}.
Therefore, we concentrate on guiding the position and scales that the generator synthesizes.

To this end, we propose a hierarchical structure of Gaussians to effectively regularize the position and scale of Gaussians, as illustrated in Fig.~\ref{fig: gaussian hierarchy}.
Firstly, we define the hierarchy level $l \in \{0, ..., L-1\}$, from coarse to fine levels, where each level contains a set of Gaussian parameters.
In detail, we establish a dependency between the Gaussian parameters of adjacent levels.
For simplification, we explain the dependency between two hierarchically adjacent Gaussians, $G^{l}$ and $G^{l-1}$, where $G^{l}$ originates from $G^{l-1}$.
We aim to model the 3D representation in a coarse-to-fine manner, by assigning coarser- and finer-level Gaussians $G^{l-1}$ and $G^{l}$ to be responsible for coarser and finer details of the 3D scene, respectively.

For the position parameter, we impose a locality constraint that bounds the position $\mu^l$ of the finer-level Gaussian $G^l$ with its corresponding $G^{l-1}$.
Specifically, we introduce the local position parameter $\hat{\mu}^l$ defined in the local coordinate system, which is centered, rotated, and scaled by the position $\mu^{l-1}$ and scale $s^{l-1}$ and quaternion $q^{l-1}$ of the coarser-level Gaussian $G^{l-1}$.
Then, the position $\mu^l$ in world space is formulated by transforming the local position $\hat{\mu}^{l}$ as follows:
\begin{equation}
    \label{eqn:locality}
    \mu^{l} = \mu^{l-1} + R^{l-1} S^{l-1} \hat{\mu}^{l}, 
\end{equation}
where $R^{l-1}$ and $S^{l-1}$ are a rotation and scaling matrix obtained from $q^{l-1}$ and $s^{l-1}$. 
This operation ensures the position of Gaussians at finer levels depends on coarser-level Gaussians, while residing near the location of coarser-level counterparts.

For the scale parameter, we enforce the scale parameter to monotonically decrease to a certain degree as its hierarchy level increases.
In detail, we define the scale of the finer-level $s^{l}$ using the relative scale difference $\hat{s}^{l}$ against the coarser-level scale $s^{l-1}$.
Additionally, we restrict this scale difference $\hat{s}^{l}$ to always be a vector of negative values.
Furthermore, we introduce the constant $\Delta s$ which further lowers the scale of the finer-level.
This process is defined as follows:
\begin{equation}
    \label{eqn:scale diff}
    s^{l} = s^{l-1} + \hat{s}^{l} + \Delta s, \quad \text{where} \quad \hat{s}^{l}, \Delta s < 0.
\end{equation}

For the other parameters, we additionally define the residual Gaussian parameters $\hat{\alpha}^l, \hat{q}^l, \hat{c}^l$ at level $l$, which are added to the Gaussian parameters of the previous level as follows:
\begin{equation}
    \label{eqn:residual params}
    \alpha^l = \alpha^{l-1} + \hat{\alpha}^l, \quad q^l = q^{l-1} + \hat{q}^l, \quad c^l = c^{l-1} + \hat{c}^l. \quad
\end{equation}

We call this hierarchical relationship between $G^{l-1}$ and Gaussians with residual parameters $\{ \hat{\mu}^l, \hat{s}^l, \hat{\alpha}^l, \hat{q}^l, \hat{c}^l \} \in \hat{G}^l$ as $\texttt{densify}(G^{l-1}, \hat{G}^{l})$, and it enables the generator to model the 3D space in a coarse-to-fine manner, where the fine-level Gaussians depict the detailed part of the coarse-level counterparts.
Importantly, it stabilizes the training of GANs by significantly reducing the possible positions of Gaussians and encourages the generator to use various scales of Gaussians, thereby boosting the generation capability to model both coarse and fine details.

\begin{figure}[t!]
     \begin{subfigure}[b]{0.78\textwidth}
        \centering
         \includegraphics[width=\textwidth]{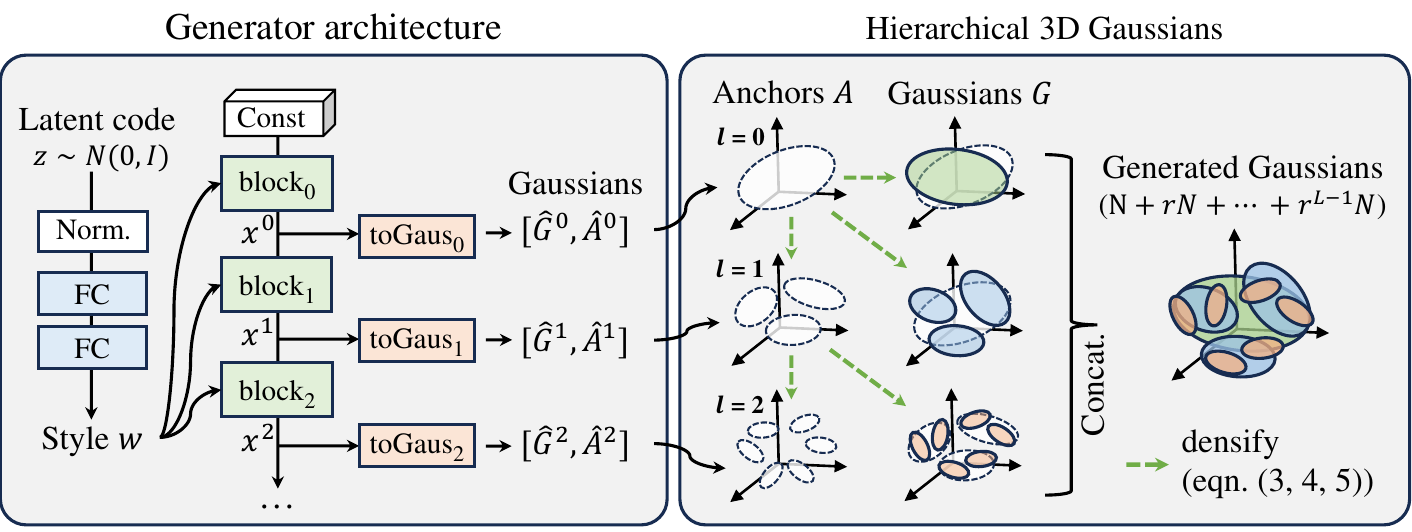}
         \caption{Illustration of generator architecture with hierarchical 3D Gaussians}
         \label{fig: generator architecture}
    \end{subfigure}
     \hspace*{0.1cm}
     \begin{subfigure}[b]{0.20\textwidth}
         \centering
         \includegraphics[width=\textwidth]{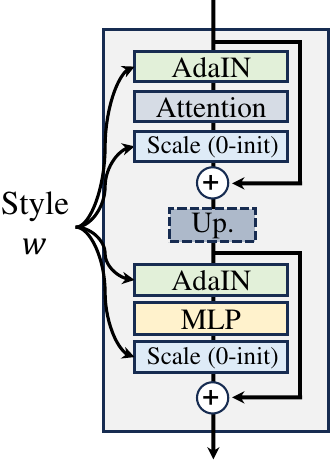} 
         \caption{$\texttt{block}$ architecture}
         \label{fig: block architecture}
    \end{subfigure}
    \centering
    \caption{
    The architecture of the generator and its block. 
    (a) Generator synthesizes the multiple-level of anchors and Gaussians, which contains the residual parameters $\hat{A}^{l}$ and $\hat{G}^l$.
    Anchors are utilized to regularize the finer-level Gaussians, while Gaussians are used for actual rendering.
    After generating these parameters, we combine them with anchors from previous level $A^{l-1}$ by $\texttt{densify}$ operation, as defined in eqn.~\ref{eqn:locality},~\ref{eqn:scale diff},~\ref{eqn:residual params}~(green arrow).
    (b) The generator consists of stacks of \texttt{block}s, each of which is a sequence of attention and MLP layers. The latent code $z$ is conditioned on the generator through AdaIN and layerscale, where the modulation and scaling parameters are derived from style code $w$. 
    }
    \label{fig: model architecture}
\end{figure}

\subsection{GSGAN; the generator architecture with hierarchical 3D Gaussians}
\label{sec:3.3_architectural_details}
In this section, we propose a generator architecture for leveraging the aforementioned hierarchical structure of Gaussians~(Fig.~\ref{fig: model architecture}).
Basically, we adopt a transformer-based architecture, composed of stacks of attention and MLP layers, which is a generally used architecture to handle unstructured 3D point cloud data~\cite{zhao2021point,guo2021pct,nichol2022point}.

First of all, we define a generator $g(z, \theta)$ as a sequence of generator blocks, where $\texttt{block}_l$ denotes the generator block at a specific level $l$.
At the coarsest level $l=0$, $\texttt{block}_0$ takes input as $N$ learnable positions $\texttt{const} \in \mathbb{R}^{N \times 3}$ and latent code $z$, where $N$ is the number initial Gaussians, then outputs the high-dimensional features $x^{0}$.
Then, for a feature of $i^{\text{th}}$ Gaussian $x^0_i$, we process the feature $x_i^0$ by the output layer $\texttt{toGaus}_0$ to obtain a Gaussian parameter $\{\mu_i^0, s_i^0, q_i^0, \alpha_i^0, c_i^0\} \in G_i^0$.
For arbitrary level $l$, $\texttt{block}_l$ takes input as feature $x_i^{l-1}$ from the previous block and latent code $z$, then outputs the feature $x_j^{l}$.
Importantly, the output block $\texttt{toGaus}_{l}$ does not directly synthesize the Gaussian parameters $G_j^l$. Instead, the intermediate output $\hat{G}_j^l$ contains the local position $\hat{\mu}_j^l$ and relative scale difference $\hat{s}_j^l$, as well as the other residual parameters $\hat{c}_j^l, \hat{\alpha}_j^l, \hat{q}_j^l$.
This intermediate output $\hat{G}_j^{l}$ is combined with the corresponding Gaussians $G_i^{l-1}$ from the previous level to finally synthesize the Gaussian $G^{l}_j$ at level $l$, following the operations in Sec.~\ref{sec:hierarchical 3D gaussian sapltting}.
This process, which establishes the hierarchical dependency between Gaussians of adjacent level $G_i^{l-1}$ and $G_j^l$, is defined as follows:
\begin{equation}
    \label{eqn: architecture hierarchy base}
    x_j^{l} = \texttt{block}_l(x_i^{l-1}, z), \quad \hat{G}_j^{l} = \texttt{toGaus}_l(x_j^{l}), \quad G_j^{l} = \texttt{densify}(G_i^{l-1}, \hat{G}_j^{l}),
\end{equation}
where $\texttt{densify}$ operation denotes the combining process of parameters in hierarchically adjacent $G_i^{l-1}$ and $\hat{G}_j^{l}$ mentioned in eqn.~\ref{eqn:locality},~\ref{eqn:scale diff},~\ref{eqn:residual params}.

As the fine-level Gaussians have smaller scales compared to the coarse-level ones, the number of Gaussians should be increased as the hierarchy increases to successfully synthesize the fine details.
Thus, we expand the number of Gaussian parameters $G_i^{l}$ to have a total of $r^{l} N$ vectors for each parameter, where $r$ is an upsampling ratio.
In other words, we define the Gaussian $G_j^{l}$ to be dependent on $G_i^{l-1}$, where $j = ri + k$ and $k \in \{ 0, 1, ..., r-1 \}$.

After synthesizing the Gaussian parameters of every level, we use all of them for generating the image~(i.e. total $(N + rN + ... + r^{L-1} N)$ Gaussians are used for rendering). For rendering, we use a tile-based rasterizer following 3D-GS~\cite{kerbl20233d}.

\paragraph{Anchor Gaussians to decompose Gaussians for regularization and rendering}
In the aforementioned architecture, Gaussians are used not only to represent a 3D scene but also to regularize their coarser-level Gaussian counterparts. This means Gaussians must be trained to precisely guide the parameters of finer-level Gaussians while simultaneously depicting the sharp details in real-world images. However, achieving both of these objectives can be challenging. For instance, we observe that the scale of Gaussians can become nearly zero along a specific axis, leading to excessively strong regularization on the position of finer-level Gaussians.
To handle this issue, we introduce an auxiliary set of Gaussians that only contributes to regularization, instead of actual rendering.

Specifically, we introduce an anchor Gaussian $A^l$ for a specific level $l$, which has identical parameterization to $G^l$.
However, this type of Gaussian is only used for regularization by deploying it as the input of $\texttt{densify}$, especially for the coarser-level Gaussian input.
Therefore, an anchor Gaussian $A_i^{l-1}$ only learns to guide the parameters of their finer-level counterpart $G_j^{l}$, so its usage relieves the effects of strong regularization caused by zero variance.
To generate it, we simply make $\texttt{toGaus}_l$ synthesize two sets of Gaussians, $\hat{G}_j^l$ and $\hat{A}_j^l$, for a given feature $x_j^l$.
This process is achieved by re-defining the eqn.~\ref{eqn: architecture hierarchy base} as follows:
\begin{equation}
    [\hat{G}_j^l, \hat{A}_j^l] = \texttt{toGaus}_l(x_j^l), \quad G_j^l = \texttt{densify}(A_i^{l-1}, \hat{G}_j^l), \quad A_j^l = \texttt{densify}(A_i^{l-1}, \hat{A}_j^l).
\end{equation}
As the Gaussians of coarsest level $l=0$ does not have their coarser-level counterparts, we define $A_i^0=\hat{A}_i^0$ and $G_i^0=\texttt{densify}(A_i^0, \hat{G}_i^0)$.

\paragraph{Architectural details}
Following the conditioning convention of previous GANs~\cite{karras2019style,karras2020analyzing}, we utilize the mapping network to modify the latent code $z$ into the style code $w$.
Then, the style code affects the synthesis process by AdaIN~\cite{huang2017arbitrary}.
As noted, the generator block is a stack of attention and MLP layers, then we replace the layer norm in attention and MLP by AdaIN, following generally used approach for transformer-based GANs~\cite{lee2021vitgan,zhang2022styleswin}.
For blocks of coarser levels, we utilize the general self-attention without positional encoding as the attention mechanism, whereas we use local attention~\cite{wang2019dynamic} for finer levels, as the interaction between $r^{l-1} N$ points is computationally demanding.
For expanding the features in generator blocks after the coarsest level, we simply use the subpixel operation~\cite{shi2016real} with skip connection and repeat Gaussians $G^l$ in $r$ times.

One important architectural design is the usage of layerscale~\cite{touvron2021going}, which is a learnable vector that adjusts the effects of the residual block by multiplying it by the output of the residual block.
Typically, it is initialized by a zero-valued vector, removing the effects of layers in the early stage of training.
We observe it is essential for stabilizing the position of Gaussians in early iterations.
In addition, we use the adaptive version of layerscale~\cite{peebles2023scalable} conditioned by latent code $z$ on every attention and MLP layer in the generator.

Also, we use the camera direction as a condition for the color layer in $\texttt{toGaus}$ to model the view-dependent characteristics and employ the background generator, which resembles the generator architecture but with reduced capacity and results in the Gaussians located within a sphere of a radius of 3, while the foreground resides in the [-1, 1] cube.
For further details, we elaborate on them in the Appendix~\ref{sec: appendix implementation details}.

\subsection{Training objectives}
Similar to previous 3D GANs~\cite{chan2022efficient}, we adopt the non-saturating adversarial loss~\cite{goodfellow2014generative} with R1 regularization~\cite{mescheder2018training}.
Formally, these objective functions are defined as follows:
\begin{equation}
    \mathcal{L}_{\text{adv}} = \mathbb{E}_{z \sim P_z, \theta \sim P_\theta}[f(d(g(z, \theta)))] + \mathbb{E}_{I_r \sim P_{I_r}}[f(-d(I_r)) + \lambda|| \nabla d(I_r) ||^2],
\end{equation}
where $f(t)=-\text{log}(1 + \text{exp}(-t))$ is a softplus function and $\lambda$ is R1 regularization strength.

We additionally guide the 3D information to the discriminator and generator by introducing contrastive learning between pose embedding obtained from the images and camera parameters.
Specifically, the discriminator has a pose branch $d_p$ that estimates the pose embedding $p_I$ from an input image.
Then, we introduce a pose encoder that consists of MLP layers and encodes the camera parameter $\theta$ into the pose embedding $p_\theta$.
Similar to previous work~\cite{jo20233d}, we utilize the contrastive objective which enhances similarity between corresponding $p_I$ and $p_\theta$.
Formally, this objective is defined as follows:
\begin{equation}
    \mathcal{L}_{\text{pose}} = - \text{log}(\frac{\text{exp}(\text{sim}(p_I, p^+_\theta) / \tau)}{\sum_{b=1}^{B}(\text{exp}(\text{sim}(p_I, p^b_\theta) / \tau))}),
\end{equation}
where $\text{sim}(\cdot, \cdot)$ is a cosine similarity and $B$ is a batch size and $p^+_\theta$ is a positive sample corresponding to a pose embedding $p_I$ from the image, and $\tau$ is a temperature scaling parameter.
For the discriminator, we calculate $\mathcal{L}_{\text{pose}}$ using real data, while using fake data for training the generator.

Furthermore, we introduce two losses to regularize the position of anchor Gaussians in the coarsest level, $\mu_A^0$.
We first regularize the averaged position of $\mu_A^0$ to be zero for encouraging the center of Gaussians residing near the origin of the world space.
Secondly, we reduce the distance between the positions of the $K$ nearest anchor Gaussians to prevent anchor Gaussians from falling apart from the others.
These two regularization losses are defined as follows:
\begin{equation}
    \mathcal{L}_{\text{center}} = \frac{1}{N} ||\sum_{j=1}^{N} \mu^0_{A, j} ||^2, \quad 
    \mathcal{L}_{\text{knn}}= \frac{1}{NK} \sum_{j=1}^{N} || \sum_{k=1}^{K} (\mu^0_{A, j}- \text{KNN}(\mu^0_{A, j}, k))||^2,
\end{equation}
where $\text{KNN}(\mu_{A, j}^0, k)$ is the position of $k^{\text{th}}$ nearest neighbor of $j^{\text{th}}$ anchor Gaussian of $\mu_A^0$.

To sum up, the final objective function $\mathcal{L}$ is as follows:
\begin{equation}
    \mathcal{L} = \mathcal{L}_{\text{adv}} + \lambda_{\text{pose}} \mathcal{L}_{\text{pose}} + \lambda_{\text{center}} \mathcal{L}_{\text{center}} + \lambda_{\text{knn}} \mathcal{L}_{\text{knn}},
\end{equation}
where $\lambda_{\text{pose}}$, $\lambda_{\text{center}}$, and $\lambda_{\text{knn}}$ are strengths of the corresponding objective function.

\section{Experiments}
\label{sec:4 experiments}

\subsection{Experimental settings}
\vspace{-0.1cm}
\label{sec:4.1 experimental settings}
Following the experimental settings of previous 3D GANs~\cite{chan2022efficient,chen2023mimic3d}, we use FFHQ~\cite{karras2019style} and AFHQ-Cat~\cite{choi2020stargan} datasets with 256$\times$256 and 512$\times$512 resolutions.
For example, we augment the datasets with the horizontal flip and additionally use adaptive data augmentation~\cite{karras2020training} for AFHQ-Cat dataset, which has a limited size.
Camera pose labels are obtained from the official repository of EG3D~\cite{chan2022efficient}, which are predicted by off-the-shelf pose estimators~\cite{deng2019accurate,catpose}.
We train the model from scratch on each dataset.
For further implementation details, please refer to Appendix~\ref{sec: appendix implementation details}.

\begin{table}
\small
\centering
\caption{
    Quantitative comparison on FFHQ and AFHQ-Cat datasets in terms of FID-50K-full and rendering time.
    We mainly compare ours with the 3D consistent models, except EG3D utilizing the 2D upsampler. 
    FID scores are taken from previous work~\cite{chen2023mimic3d}.
    Rendering time is measured on a single RTX A6000 GPU. 
    $^\ast$The rendering time of EG3D at 512 resolution consists of the time for volume rendering at 128 resolution and 2D upsampling operations.
}
\begin{tabular}{c|c|cc|cc|cc}
\toprule
            &  3D              & \multicolumn{2}{c|}{FFHQ}  &   \multicolumn{2}{c|}{AFHQ-Cat} & \multicolumn{2}{c}{Rendering time~(ms)} \\
 Methods    &  consistency  & 256$\times$256 & 512$\times$512 & 256$\times$256 & 512$\times$512 & 256$\times$256 & 512$\times$512 \\
\midrule
EG3D~\cite{chan2022efficient}&  &  4.80 &   4.70 &   3.41 &   2.72 &  - &  15.5$^\ast$     \\
\midrule
GRAM~\cite{deng2022gram}        & \checkmark &  13.8 &   - &   13.4 &   - &  - & -     \\
GMPI~\cite{zhao2022generative}        & \checkmark &  11.4 &   8.29 &   - &   7.67 &   - & -    \\
EpiGRAF~\cite{skorokhodov2022epigraf}     & \checkmark &   9.71 &   9.92 &   6.93 &   - &   - & -    \\
Voxgraf~\cite{schwarz2022voxgraf}     & \checkmark &   9.60 &   - &   9.60 &   - &  - & -     \\
GRAM-HD~\cite{xiang2023gram}     & \checkmark &   10.4 &   - &   - &   7.67 &  173.0 & 197.9     \\
Mimic3D~\cite{chen2023mimic3d}    & \checkmark &   \textbf{5.14} &   \textbf{5.37} &   \underline{4.14} &   \underline{4.29} & 106.8 & 402.1      \\
\midrule
GSGAN (Ours)        & \checkmark &   \underline{6.59} &   \underline{5.60} &   \textbf{3.43} &   \textbf{3.79} &   \textbf{2.7} & \textbf{3.0}    \\
\bottomrule
\end{tabular}
\vspace{-0.4cm}
\label{table: main}
\end{table}

\subsection{Experimental results}
\vspace{-0.1cm}
\paragraph{Quantitative results}
We mainly compare the proposed method with previous 3D consistent GANs, in terms of FID. These methods have strict 3D consistency, which directly renders the high-resolution images from 3D representation without any 2D upsampling operations.
As reported in Tab.~\ref{table: main}, we validate that the proposed method surpasses most of the previous methods and also achieves comparable generation capability compared to the state-of-the-art, Mimic3D.
Especially for AFHQ-Cat dataset, we achieve a much lower FID, even comparable to the non-3D consistent baseline, EG3D.
Next, we evaluate the rendering speed enhancement of the proposed method compared to baseline methods.
To compute it, we first synthesize a 3D representation of each model and measure the processing time of rendering.
As reported, ours shows significantly faster speeds compared to the baseline methods, achieving more than 100 times faster rendering than Mimic3D in 512$\times$512 resolution.
Moreover, it is even faster than non-3D consistent baseline EG3D, which is the model that exploits 2D upsampling operation for reducing the efficient rendering.
In addition, one important point is that rendering time is almost identical regardless of the image resolution, suggesting that the proposed method can be more effective at higher resolutions.
Additionally, the training time of the proposed methods is 28 RTX A6000 days, while the state-of-the-art Mimic3D requires 64 A100 days on the FFHQ-512 dataset.
This notable gap in training cost implies that ours can achieve comparable generation capability efficiently in both rendering and training speed.

\begin{figure}[t!]
    \centering
    \includegraphics[width=1.0\columnwidth]{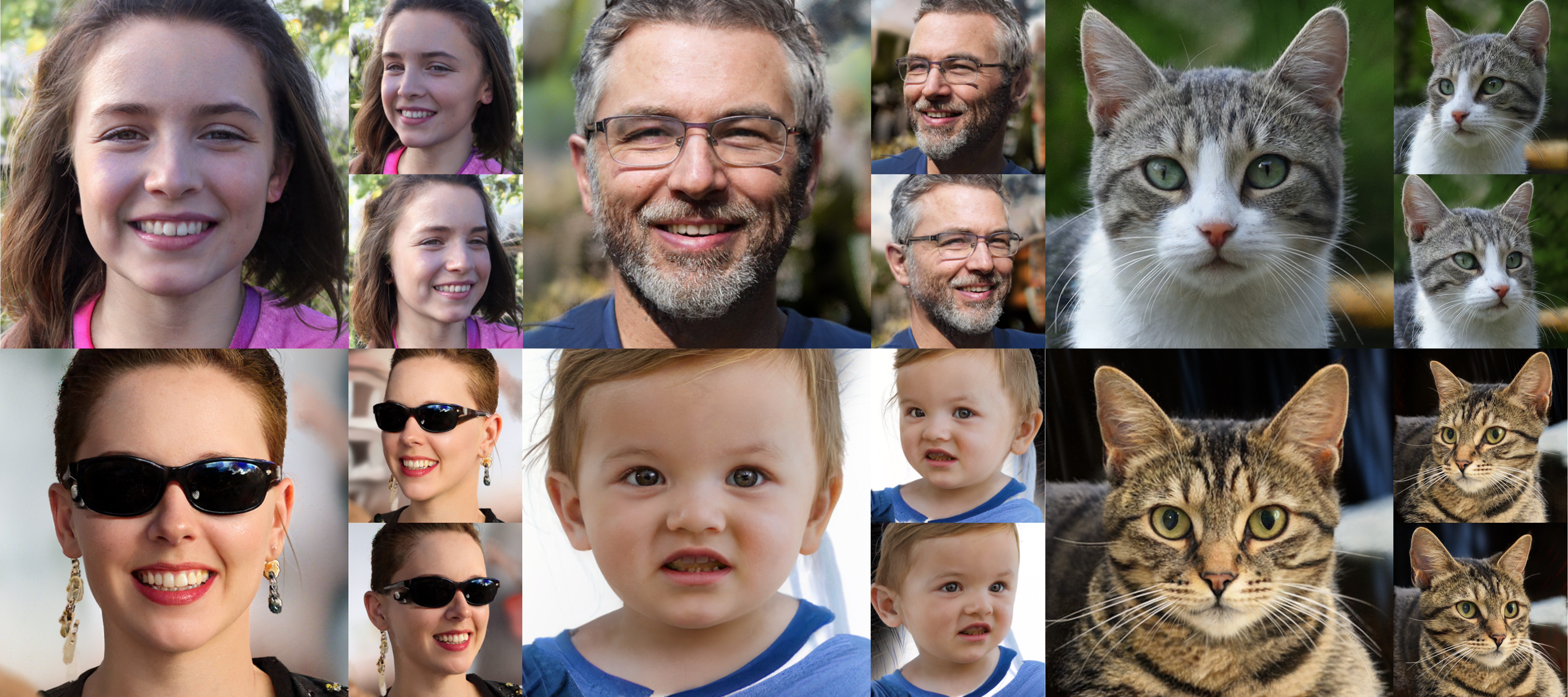}
    \caption{
    Qualitative results of the proposed method with truncation psi~($\psi=0.7$).}
    \vspace{-0.2cm}
    \label{fig: examples ours only}
\end{figure}

\begin{figure}[t!]
    \centering
    \includegraphics[width=1.0\columnwidth]{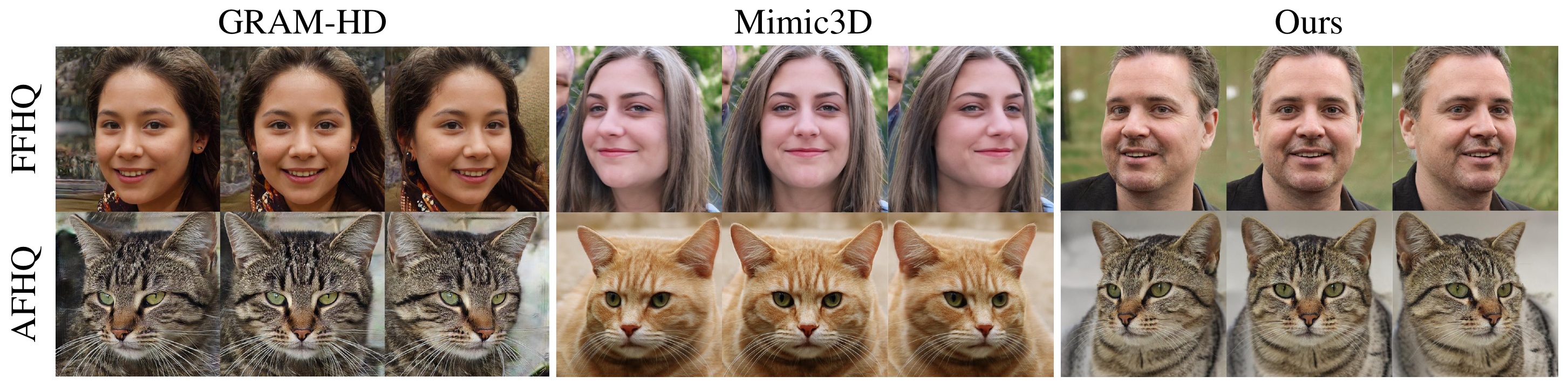}
    \caption{Qualitative comparison with 3D consistent methods with truncation psi~($\psi=0.7$).}
    \vspace{-0.3cm}
    \label{fig: qualitative vs. baseline}
\end{figure}

\begin{figure}
    \centering
    \includegraphics[width=\textwidth]{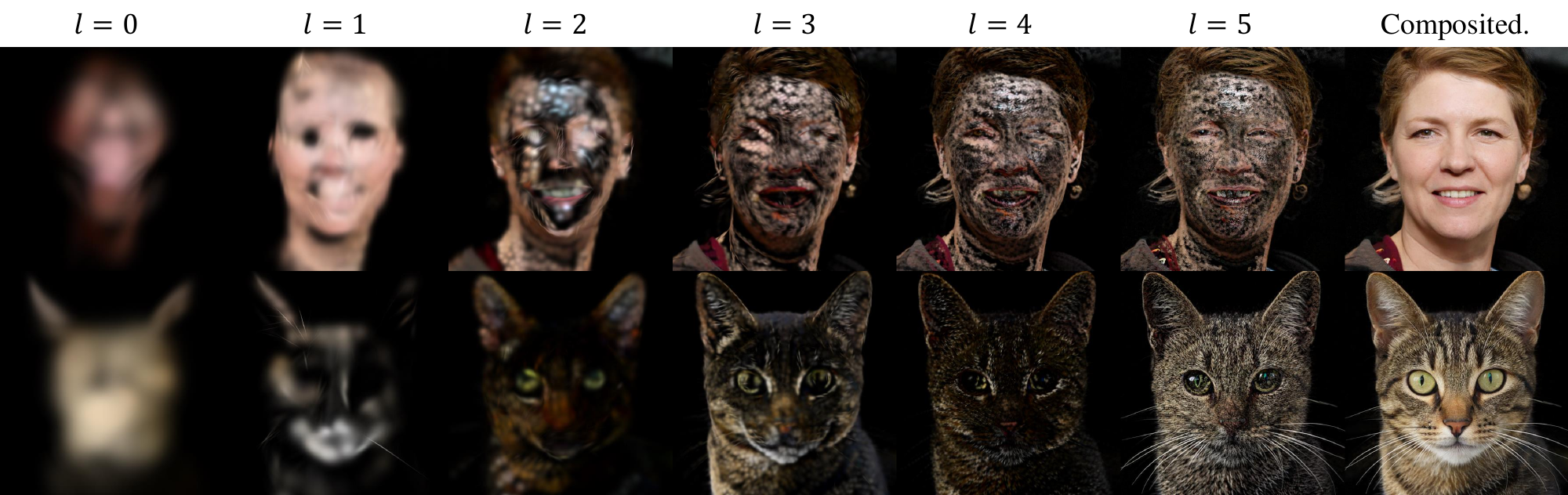}
    \caption{Level-by-level visualization of synthesized Gaussians.}
    \label{fig: example per-level}
\end{figure}

\paragraph{Qualitative results}
We present examples of generated images from the proposed method and the most recent 3D consistent GANs~\cite{chen2023mimic3d,xiang2023gram}, in Fig.~\ref{fig: qualitative vs. baseline}.
In both datasets, we observe that the proposed method successfully synthesizes the multi-view consistent images, validating its capability for synthesizing the realistic 3D scene.
Also, we validate ours can generate both coarse and fine details such as coarse details of skin in human facial images and fine details of fur in cat images.

\paragraph{Level-by-level visualization}
To further understand how the synthesized Gaussians work, we visualize the Gaussians from an individual level. As depicted in Fig.~\ref{fig: example per-level}, we observe that Gaussians capture the image components from overall structure to fine details as the level increases.

\begin{table}
    \vspace{-0.7cm}
  \begin{minipage}[h]{.55\linewidth}
    \small
    \centering
        \caption{
        Comparison of 3D consistency. We compare ours with the most recent 3D consistent GANs~\cite{xiang2023gram,chen2023mimic3d}. 
        Faces are segmented by off-the-shelf model~\cite{yu2018bisenet} to remove effects of background.
        }
        \begin{tabular}{c|cc|cc}
        \toprule
            & \multicolumn{2}{c}{FFHQ-256} & \multicolumn{2}{c}{FFHQ-512} \\
         & PSNR & SSIM & PSNR & SSIM \\
        \midrule
        GRAM-HD     &   34.45 &   0.9648 &   32.15 &  0.9244       \\
        Mimic3D    &   40.36 &   \textbf{0.9927} &   35.95 &  \textbf{0.9822}         \\
        \bottomrule
        GSGAN        &   \textbf{41.69} &   0.9883 &   \textbf{37.85} &  0.9695   \\
        \end{tabular}
        \label{tab: 3d consistency}
  \end{minipage}
  \hfill
  \begin{minipage}[h]{.4\linewidth}
  \vspace{0.4cm}
    \centering
    \includegraphics[width=1.0\linewidth]{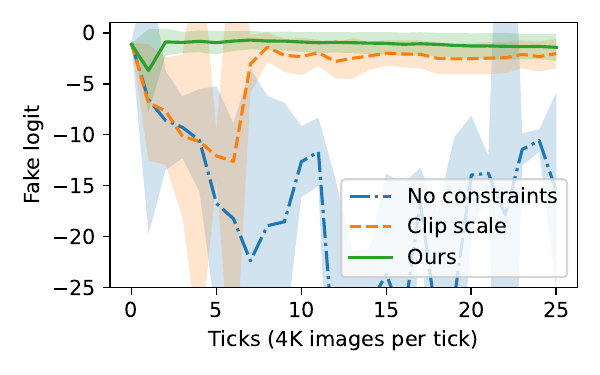}
    \captionof{figure}{Fake logits of ablated models at the early stage of training.}
    \label{fig: instability early stage}
  \end{minipage}
  \vspace{-0.6cm}
\end{table}
\paragraph{Comparison with 3D consistency}
As a 3D generative model, it is important to maintain 3D consistency across different views.
To validate the 3D consistency of the synthesized 3D model, we measure how well the generated 3D scene is reconstructed by a surface estimation model, following previous works~\cite{chen2023mimic3d,xiang2023gram}. Specifically, we fit the surface estimation model to a generated multi-view image and compute the reconstruction error of the multi-view images used for training.
For surface estimation, we use Neus2~\cite{wang2023neus2}, applying a facial segmentation mask estimated from an off-the-shelf network~\cite{yu2018bisenet} to eliminate the effects of the background. Furthermore, we utilize only the foreground generator for our method.
As shown in Tab.~\ref{tab: 3d consistency}, the proposed method significantly outperforms GRAM-HD while achieving performance comparable to Mimic3D. This experiment suggests that our approach generates 3D-consistent results by leveraging the explicit 3D representation provided by 3D Gaussians.

\paragraph{Training stability at the early stage of training}
We conduct an experiment to assess the effect of the proposed method on training stability, particularly at the early stage of training, by observing the fake logit of the discriminator.
As shown in Fig.~\ref{fig: instability early stage}, we observe that the model without any constraints exhibits rapid divergence, accompanied by a markedly low fake logit, indicating that the discriminator already distinguishes between real and fake data perfectly.
When applying a minimal constraint that limits the scale to its predefined maximum, the model does not diverge but still suffers from instability, as evidenced by a large standard deviation of the logit.
In contrast, ours demonstrates stable training compared to other models.
Note that, the standard deviation is visualized by 1-$\sigma$.

\begin{table}

  \vspace{-0.4cm}
  \begin{minipage}[h]{.4\linewidth}
    \centering
        \caption{
        Ablation study on FFHQ-256. 
        ``Clipping scale" means the model with the maximum limit of scale. 
        }
        \begin{tabular}{l|c}
        \toprule
         & FID \\
        \midrule
        No constraints                  &  300$\sim$ \\
        \midrule
        + Clipping scale                     & 95.97  \\
        + Position reg.~(eqn.~\ref{eqn:locality})       & 17.65  \\
        + Scale reg.~(eqn.~\ref{eqn:scale diff})          & 13.80  \\
        + Background generator       &  12.61 \\
        + Anchor Gaussian        &   6.59 \\
        \bottomrule
        \end{tabular}
        \label{tab: ablation}
  \end{minipage}
  \hfill
  \begin{minipage}[h]{.55\linewidth}
  \vspace{0.4cm}
    \centering
    \includegraphics[width=1.0\linewidth]{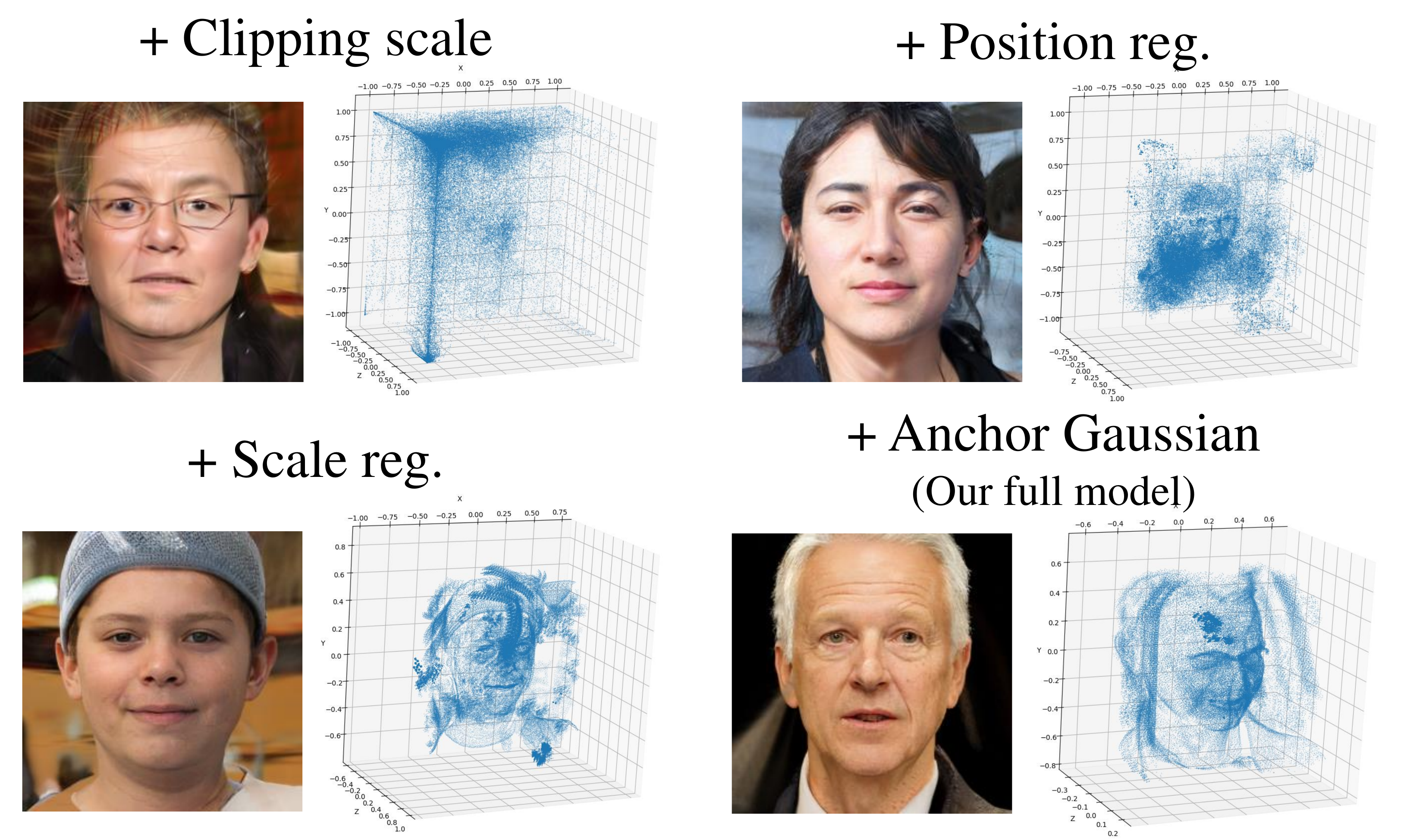}
    \captionof{figure}{Visual comparison with ablated models. Points are visualized by the position of Gaussians.}
    \label{fig: ablation point clouds}
  \end{minipage}
  \vspace{-0.5cm}
\end{table}
\paragraph{Ablation study}
We perform an ablation study on the proposed components, particularly focusing on the proposed hierarchical architecture.
When ablating the regularizations, we keep the residual representation of other parameters, ($q, \alpha, c$).
In the absence of any constraints, the model exhibited an FID score exceeding 300, signifying a failure to converge, in Tab.~\ref{tab: ablation}.
With a minimal constraint that clips scales, the model does not diverge but significantly suffers from its low generation capability.
As we gradually attach the proposed components, we observe enhancements in FID, showing the effects of the regularization of position and scale and the introduction of anchor Gaussians.

In Fig.~\ref{fig: ablation point clouds}, we provide visualizations of synthesized Gaussian positions.
Initially, a simple clipping of the scale allow the model to synthesize images, but it leads to visual artifacts due to elongated Gaussians with large scales. 
Additionally, many Gaussians remain outside the scene, likely due to their tendency not to overlap, particularly when large-scale Gaussians exist.
After applying position regularization, Gaussians become more densely located, although visual artifacts and unused Gaussians persist.
Upon introducing scale regularization, the synthesized images no longer exhibit such artifacts, but the model struggles to capture precise geometries, as evidenced by the point cloud visualizations.
Finally, the model incorporating anchor Gaussians successfully synthesizes realistic images while accurately estimating Gaussian positions. 
Note that, we clip the Gaussian positions to exist within the range of [-1, 1] in the cube.

\paragraph{Additional visualizations}
We additionally provide 1) additional generated examples, 2) examples with densely changed camera positions, 3) latent interpolation and w+ inversion, 4) visualization of anchor Gaussians, and 5) effects of background generator in Appendix~\ref{sec: appendix}, and code implementation with additional videos in supplementary zip file, so please refer to them.
\section{Broader Impact and Limitations}
\vspace{-0.1cm}
\paragraph{Broader Impact}
\label{sec: broader impact}
The proposed method follows the negative social impact of previous 3D generative models.
For example, it may be used for synthesizing fake news or deepfake.
Furthermore, since we boost the rendering speed of 3D GANs, $\times$100 faster than previous methods, it can encourage the generation of them by reducing computation cost for rendering the image.

\vspace{-0.1cm}
\paragraph{Limitations}
\label{sec: limitation}
Different from the 3D-GS, which adaptively removes and introduces Gaussians by densification, the proposed method synthesizes a fixed number of Gaussians.
This lack of adaptivity in the number of Gaussians remains a limitation, as the number of Gaussians can differ depending on the scene it makes.
Also, the scale in hierarchical Gaussian representation is somewhat dependent on the hyperparameter such as $\Delta s$.
These factors require adjustment of hyperparameters and can affect the performance of the generator.

\section{Conclusion}
\label{sec: conclusion}
In this paper, we exploit the 3D Gaussian representation in the domain of 3D GANs, leveraging its fast rendering speed with explicit 3D representation.
To relieve the absence of proper initialization and densification process of 3D-GS, we propose hierarchical Gaussian representation which effectively regularizes the position and scale of generated Gaussians.
We validate the proposed technique to stabilize the training of the generator with 3D Gaussians and encourage the model to learn the precise geometry of 3D scene, achieving almost ~$\times$100 faster rendering speed with a comparable 3D generation capability.

\section{Acknowledgements}
This work was supported in part by MSIT\&KNPA/KIPoT (Police Lab 2.0, No. 210121M06), MSIT/IITP (No. 2022-0-00680, 2020-0-01821, 2019-0-00421, RS-2024-00459618, RS-2024-00360227, RS-2024-00437102, RS-2024-00437633), and MSIT/NRF (No. RS-2024-00357729).

\newpage
{\small
\bibliographystyle{unsrtnat}
\bibliography{bibliography}

\begin{thebibliography}{60}
\providecommand{\natexlab}[1]{#1}
\providecommand{\url}[1]{\texttt{#1}}
\expandafter\ifx\csname urlstyle\endcsname\relax
  \providecommand{\doi}[1]{doi: #1}\else
  \providecommand{\doi}{doi: \begingroup \urlstyle{rm}\Url}\fi

\bibitem[Jain et~al.(2022)Jain, Mildenhall, Barron, Abbeel, and Poole]{jain2022zero}
Ajay Jain, Ben Mildenhall, Jonathan~T Barron, Pieter Abbeel, and Ben Poole.
\newblock Zero-shot text-guided object generation with dream fields.
\newblock In \emph{Proceedings of the IEEE/CVF conference on computer vision and pattern recognition}, pages 867--876, 2022.

\bibitem[Poole et~al.(2022)Poole, Jain, Barron, and Mildenhall]{poole2022dreamfusion}
Ben Poole, Ajay Jain, Jonathan~T Barron, and Ben Mildenhall.
\newblock Dreamfusion: Text-to-3d using 2d diffusion.
\newblock \emph{arXiv preprint arXiv:2209.14988}, 2022.

\bibitem[Tang et~al.(2023{\natexlab{a}})Tang, Ren, Zhou, Liu, and Zeng]{tang2023dreamgaussian}
Jiaxiang Tang, Jiawei Ren, Hang Zhou, Ziwei Liu, and Gang Zeng.
\newblock Dreamgaussian: Generative gaussian splatting for efficient 3d content creation.
\newblock \emph{arXiv preprint arXiv:2309.16653}, 2023{\natexlab{a}}.

\bibitem[Wang et~al.(2023{\natexlab{a}})Wang, Lu, Wang, Bao, Li, Su, and Zhu]{wang2023prolificdreamer}
Zhengyi Wang, Cheng Lu, Yikai Wang, Fan Bao, Chongxuan Li, Hang Su, and Jun Zhu.
\newblock Prolificdreamer: High-fidelity and diverse text-to-3d generation with variational score distillation.
\newblock In \emph{Thirty-seventh Conference on Neural Information Processing Systems}, 2023{\natexlab{a}}.

\bibitem[Lin et~al.(2023)Lin, Gao, Tang, Takikawa, Zeng, Huang, Kreis, Fidler, Liu, and Lin]{lin2023magic3d}
Chen-Hsuan Lin, Jun Gao, Luming Tang, Towaki Takikawa, Xiaohui Zeng, Xun Huang, Karsten Kreis, Sanja Fidler, Ming-Yu Liu, and Tsung-Yi Lin.
\newblock Magic3d: High-resolution text-to-3d content creation.
\newblock In \emph{Proceedings of the IEEE/CVF Conference on Computer Vision and Pattern Recognition}, pages 300--309, 2023.

\bibitem[Wang et~al.(2023{\natexlab{b}})Wang, Du, Li, Yeh, and Shakhnarovich]{wang2023score}
Haochen Wang, Xiaodan Du, Jiahao Li, Raymond~A Yeh, and Greg Shakhnarovich.
\newblock Score jacobian chaining: Lifting pretrained 2d diffusion models for 3d generation.
\newblock In \emph{Proceedings of the IEEE/CVF Conference on Computer Vision and Pattern Recognition}, pages 12619--12629, 2023{\natexlab{b}}.

\bibitem[Chen et~al.(2023{\natexlab{a}})Chen, Chen, Jiao, and Jia]{chen2023fantasia3d}
Rui Chen, Yongwei Chen, Ningxin Jiao, and Kui Jia.
\newblock Fantasia3d: Disentangling geometry and appearance for high-quality text-to-3d content creation.
\newblock In \emph{Proceedings of the IEEE/CVF International Conference on Computer Vision}, pages 22246--22256, 2023{\natexlab{a}}.

\bibitem[Raj et~al.(2023)Raj, Kaza, Poole, Niemeyer, Ruiz, Mildenhall, Zada, Aberman, Rubinstein, Barron, et~al.]{raj2023dreambooth3d}
Amit Raj, Srinivas Kaza, Ben Poole, Michael Niemeyer, Nataniel Ruiz, Ben Mildenhall, Shiran Zada, Kfir Aberman, Michael Rubinstein, Jonathan Barron, et~al.
\newblock Dreambooth3d: Subject-driven text-to-3d generation.
\newblock In \emph{Proceedings of the IEEE/CVF International Conference on Computer Vision}, pages 2349--2359, 2023.

\bibitem[Tang et~al.(2023{\natexlab{b}})Tang, Wang, Zhang, Zhang, Yi, Ma, and Chen]{tang2023make}
Junshu Tang, Tengfei Wang, Bo~Zhang, Ting Zhang, Ran Yi, Lizhuang Ma, and Dong Chen.
\newblock Make-it-3d: High-fidelity 3d creation from a single image with diffusion prior.
\newblock In \emph{Proceedings of the IEEE/CVF International Conference on Computer Vision}, pages 22819--22829, 2023{\natexlab{b}}.

\bibitem[Melas-Kyriazi et~al.(2023)Melas-Kyriazi, Laina, Rupprecht, and Vedaldi]{melas2023realfusion}
Luke Melas-Kyriazi, Iro Laina, Christian Rupprecht, and Andrea Vedaldi.
\newblock Realfusion: 360deg reconstruction of any object from a single image.
\newblock In \emph{Proceedings of the IEEE/CVF conference on computer vision and pattern recognition}, pages 8446--8455, 2023.

\bibitem[Chan et~al.(2021)Chan, Monteiro, Kellnhofer, Wu, and Wetzstein]{chan2021pi}
Eric~R Chan, Marco Monteiro, Petr Kellnhofer, Jiajun Wu, and Gordon Wetzstein.
\newblock pi-gan: Periodic implicit generative adversarial networks for 3d-aware image synthesis.
\newblock In \emph{Proceedings of the IEEE/CVF conference on computer vision and pattern recognition}, pages 5799--5809, 2021.

\bibitem[Chan et~al.(2022)Chan, Lin, Chan, Nagano, Pan, De~Mello, Gallo, Guibas, Tremblay, Khamis, et~al.]{chan2022efficient}
Eric~R Chan, Connor~Z Lin, Matthew~A Chan, Koki Nagano, Boxiao Pan, Shalini De~Mello, Orazio Gallo, Leonidas~J Guibas, Jonathan Tremblay, Sameh Khamis, et~al.
\newblock Efficient geometry-aware 3d generative adversarial networks.
\newblock In \emph{Proceedings of the IEEE/CVF conference on computer vision and pattern recognition}, pages 16123--16133, 2022.

\bibitem[Deng et~al.(2022)Deng, Yang, Xiang, and Tong]{deng2022gram}
Yu~Deng, Jiaolong Yang, Jianfeng Xiang, and Xin Tong.
\newblock Gram: Generative radiance manifolds for 3d-aware image generation.
\newblock In \emph{Proceedings of the IEEE/CVF Conference on Computer Vision and Pattern Recognition}, pages 10673--10683, 2022.

\bibitem[Gu et~al.(2021)Gu, Liu, Wang, and Theobalt]{gu2021stylenerf}
Jiatao Gu, Lingjie Liu, Peng Wang, and Christian Theobalt.
\newblock Stylenerf: A style-based 3d-aware generator for high-resolution image synthesis.
\newblock \emph{arXiv preprint arXiv:2110.08985}, 2021.

\bibitem[Skorokhodov et~al.(2022)Skorokhodov, Tulyakov, Wang, and Wonka]{skorokhodov2022epigraf}
Ivan Skorokhodov, Sergey Tulyakov, Yiqun Wang, and Peter Wonka.
\newblock Epigraf: Rethinking training of 3d gans.
\newblock \emph{Advances in Neural Information Processing Systems}, 35:\penalty0 24487--24501, 2022.

\bibitem[Nguyen-Phuoc et~al.(2019)Nguyen-Phuoc, Li, Theis, Richardt, and Yang]{nguyen2019hologan}
Thu Nguyen-Phuoc, Chuan Li, Lucas Theis, Christian Richardt, and Yong-Liang Yang.
\newblock Hologan: Unsupervised learning of 3d representations from natural images.
\newblock In \emph{Proceedings of the IEEE/CVF International Conference on Computer Vision}, pages 7588--7597, 2019.

\bibitem[Kajiya and Von~Herzen(1984)]{kajiya1984ray}
James~T Kajiya and Brian~P Von~Herzen.
\newblock Ray tracing volume densities.
\newblock \emph{ACM SIGGRAPH computer graphics}, 18\penalty0 (3):\penalty0 165--174, 1984.

\bibitem[Mildenhall et~al.(2021)Mildenhall, Srinivasan, Tancik, Barron, Ramamoorthi, and Ng]{mildenhall2021nerf}
Ben Mildenhall, Pratul~P Srinivasan, Matthew Tancik, Jonathan~T Barron, Ravi Ramamoorthi, and Ren Ng.
\newblock Nerf: Representing scenes as neural radiance fields for view synthesis.
\newblock \emph{Communications of the ACM}, 65\penalty0 (1):\penalty0 99--106, 2021.

\bibitem[Barron et~al.(2021)Barron, Mildenhall, Tancik, Hedman, Martin-Brualla, and Srinivasan]{barron2021mip}
Jonathan~T Barron, Ben Mildenhall, Matthew Tancik, Peter Hedman, Ricardo Martin-Brualla, and Pratul~P Srinivasan.
\newblock Mip-nerf: A multiscale representation for anti-aliasing neural radiance fields.
\newblock In \emph{Proceedings of the IEEE/CVF International Conference on Computer Vision}, pages 5855--5864, 2021.

\bibitem[Barron et~al.(2022)Barron, Mildenhall, Verbin, Srinivasan, and Hedman]{barron2022mip}
Jonathan~T Barron, Ben Mildenhall, Dor Verbin, Pratul~P Srinivasan, and Peter Hedman.
\newblock Mip-nerf 360: Unbounded anti-aliased neural radiance fields.
\newblock In \emph{Proceedings of the IEEE/CVF Conference on Computer Vision and Pattern Recognition}, pages 5470--5479, 2022.

\bibitem[M{\"u}ller et~al.(2022)M{\"u}ller, Evans, Schied, and Keller]{muller2022instant}
Thomas M{\"u}ller, Alex Evans, Christoph Schied, and Alexander Keller.
\newblock Instant neural graphics primitives with a multiresolution hash encoding.
\newblock \emph{ACM transactions on graphics (TOG)}, 41\penalty0 (4):\penalty0 1--15, 2022.

\bibitem[Kerbl et~al.(2023)Kerbl, Kopanas, Leimk{\"u}hler, and Drettakis]{kerbl20233d}
Bernhard Kerbl, Georgios Kopanas, Thomas Leimk{\"u}hler, and George Drettakis.
\newblock 3d gaussian splatting for real-time radiance field rendering.
\newblock \emph{ACM Transactions on Graphics}, 42\penalty0 (4):\penalty0 1--14, 2023.

\bibitem[Snavely et~al.(2006)Snavely, Seitz, and Szeliski]{snavely2006photo}
Noah Snavely, Steven~M Seitz, and Richard Szeliski.
\newblock Photo tourism: exploring photo collections in 3d.
\newblock In \emph{ACM siggraph 2006 papers}, pages 835--846. 2006.

\bibitem[Schwarz et~al.(2022)Schwarz, Sauer, Niemeyer, Liao, and Geiger]{schwarz2022voxgraf}
Katja Schwarz, Axel Sauer, Michael Niemeyer, Yiyi Liao, and Andreas Geiger.
\newblock Voxgraf: Fast 3d-aware image synthesis with sparse voxel grids.
\newblock \emph{Advances in Neural Information Processing Systems}, 35:\penalty0 33999--34011, 2022.

\bibitem[Chen et~al.(2023{\natexlab{b}})Chen, Deng, and Wang]{chen2023mimic3d}
Xingyu Chen, Yu~Deng, and Baoyuan Wang.
\newblock Mimic3d: Thriving 3d-aware gans via 3d-to-2d imitation.
\newblock In \emph{2023 IEEE/CVF International Conference on Computer Vision (ICCV)}, pages 2338--2348. IEEE Computer Society, 2023{\natexlab{b}}.

\bibitem[Trevithick et~al.(2024)Trevithick, Chan, Takikawa, Iqbal, De~Mello, Chandraker, Ramamoorthi, and Nagano]{trevithick2024you}
Alex Trevithick, Matthew Chan, Towaki Takikawa, Umar Iqbal, Shalini De~Mello, Manmohan Chandraker, Ravi Ramamoorthi, and Koki Nagano.
\newblock What you see is what you gan: Rendering every pixel for high-fidelity geometry in 3d gans.
\newblock In \emph{Proceedings of the IEEE/CVF Conference on Computer Vision and Pattern Recognition}, pages 22765--22775, 2024.

\bibitem[Qian et~al.(2023)Qian, Kirschstein, Schoneveld, Davoli, Giebenhain, and Nie{\ss}ner]{qian2023gaussianavatars}
Shenhan Qian, Tobias Kirschstein, Liam Schoneveld, Davide Davoli, Simon Giebenhain, and Matthias Nie{\ss}ner.
\newblock Gaussianavatars: Photorealistic head avatars with rigged 3d gaussians.
\newblock \emph{arXiv preprint arXiv:2312.02069}, 2023.

\bibitem[Li et~al.(2023)Li, Zheng, Wang, and Liu]{li2023animatable}
Zhe Li, Zerong Zheng, Lizhen Wang, and Yebin Liu.
\newblock Animatable gaussians: Learning pose-dependent gaussian maps for high-fidelity human avatar modeling.
\newblock \emph{arXiv preprint arXiv:2311.16096}, 2023.

\bibitem[Xu et~al.(2023)Xu, Chen, Li, Zhang, Wang, Zheng, and Liu]{xu2023gaussian}
Yuelang Xu, Benwang Chen, Zhe Li, Hongwen Zhang, Lizhen Wang, Zerong Zheng, and Yebin Liu.
\newblock Gaussian head avatar: Ultra high-fidelity head avatar via dynamic gaussians.
\newblock \emph{arXiv preprint arXiv:2312.03029}, 2023.

\bibitem[Zhou et~al.(2024)Zhou, Ma, Fan, and Yang]{zhou2024headstudio}
Zhenglin Zhou, Fan Ma, Hehe Fan, and Yi~Yang.
\newblock Headstudio: Text to animatable head avatars with 3d gaussian splatting.
\newblock \emph{arXiv preprint arXiv:2402.06149}, 2024.

\bibitem[Abdal et~al.(2024)Abdal, Yifan, Shi, Xu, Po, Kuang, Chen, Yeung, and Wetzstein]{abdal2024gaussian}
Rameen Abdal, Wang Yifan, Zifan Shi, Yinghao Xu, Ryan Po, Zhengfei Kuang, Qifeng Chen, Dit-Yan Yeung, and Gordon Wetzstein.
\newblock Gaussian shell maps for efficient 3d human generation.
\newblock In \emph{Proceedings of the IEEE/CVF Conference on Computer Vision and Pattern Recognition}, pages 9441--9451, 2024.

\bibitem[Loper et~al.(2023)Loper, Mahmood, Romero, Pons-Moll, and Black]{loper2023smpl}
Matthew Loper, Naureen Mahmood, Javier Romero, Gerard Pons-Moll, and Michael~J Black.
\newblock Smpl: A skinned multi-person linear model.
\newblock In \emph{Seminal Graphics Papers: Pushing the Boundaries, Volume 2}, pages 851--866. 2023.

\bibitem[Kirschstein et~al.(2024)Kirschstein, Giebenhain, Tang, Georgopoulos, and Nie{\ss}ner]{kirschstein2024gghead}
Tobias Kirschstein, Simon Giebenhain, Jiapeng Tang, Markos Georgopoulos, and Matthias Nie{\ss}ner.
\newblock Gghead: Fast and generalizable 3d gaussian heads.
\newblock \emph{arXiv preprint arXiv:2406.09377}, 2024.

\bibitem[Li et~al.(2017)Li, Bolkart, Black, Li, and Romero]{FLAME:SiggraphAsia2017}
Tianye Li, Timo Bolkart, Michael.~J. Black, Hao Li, and Javier Romero.
\newblock Learning a model of facial shape and expression from {4D} scans.
\newblock \emph{ACM Transactions on Graphics, (Proc. SIGGRAPH Asia)}, 36\penalty0 (6):\penalty0 194:1--194:17, 2017.
\newblock URL \url{https://doi.org/10.1145/3130800.3130813}.

\bibitem[Karras et~al.(2019)Karras, Laine, and Aila]{karras2019style}
Tero Karras, Samuli Laine, and Timo Aila.
\newblock A style-based generator architecture for generative adversarial networks.
\newblock In \emph{Proceedings of the IEEE/CVF conference on computer vision and pattern recognition}, pages 4401--4410, 2019.

\bibitem[Jo et~al.(2023)Jo, Jin, Choo, Lee, and Cho]{jo20233d}
Kyungmin Jo, Wonjoon Jin, Jaegul Choo, Hyunjoon Lee, and Sunghyun Cho.
\newblock 3d-aware generative model for improved side-view image synthesis.
\newblock In \emph{Proceedings of the IEEE/CVF International Conference on Computer Vision}, pages 22862--22872, 2023.

\bibitem[Zhao et~al.(2021)Zhao, Jiang, Jia, Torr, and Koltun]{zhao2021point}
Hengshuang Zhao, Li~Jiang, Jiaya Jia, Philip~HS Torr, and Vladlen Koltun.
\newblock Point transformer.
\newblock In \emph{Proceedings of the IEEE/CVF international conference on computer vision}, pages 16259--16268, 2021.

\bibitem[Guo et~al.(2021)Guo, Cai, Liu, Mu, Martin, and Hu]{guo2021pct}
Meng-Hao Guo, Jun-Xiong Cai, Zheng-Ning Liu, Tai-Jiang Mu, Ralph~R Martin, and Shi-Min Hu.
\newblock Pct: Point cloud transformer.
\newblock \emph{Computational Visual Media}, 7:\penalty0 187--199, 2021.

\bibitem[Nichol et~al.(2022)Nichol, Jun, Dhariwal, Mishkin, and Chen]{nichol2022point}
Alex Nichol, Heewoo Jun, Prafulla Dhariwal, Pamela Mishkin, and Mark Chen.
\newblock Point-e: A system for generating 3d point clouds from complex prompts.
\newblock \emph{arXiv preprint arXiv:2212.08751}, 2022.

\bibitem[Karras et~al.(2020{\natexlab{a}})Karras, Laine, Aittala, Hellsten, Lehtinen, and Aila]{karras2020analyzing}
Tero Karras, Samuli Laine, Miika Aittala, Janne Hellsten, Jaakko Lehtinen, and Timo Aila.
\newblock Analyzing and improving the image quality of stylegan.
\newblock In \emph{Proceedings of the IEEE/CVF conference on computer vision and pattern recognition}, pages 8110--8119, 2020{\natexlab{a}}.

\bibitem[Huang and Belongie(2017)]{huang2017arbitrary}
Xun Huang and Serge Belongie.
\newblock Arbitrary style transfer in real-time with adaptive instance normalization.
\newblock In \emph{Proceedings of the IEEE international conference on computer vision}, pages 1501--1510, 2017.

\bibitem[Lee et~al.(2021)Lee, Chang, Jiang, Zhang, Tu, and Liu]{lee2021vitgan}
Kwonjoon Lee, Huiwen Chang, Lu~Jiang, Han Zhang, Zhuowen Tu, and Ce~Liu.
\newblock Vitgan: Training gans with vision transformers.
\newblock \emph{arXiv preprint arXiv:2107.04589}, 2021.

\bibitem[Zhang et~al.(2022)Zhang, Gu, Zhang, Bao, Chen, Wen, Wang, and Guo]{zhang2022styleswin}
Bowen Zhang, Shuyang Gu, Bo~Zhang, Jianmin Bao, Dong Chen, Fang Wen, Yong Wang, and Baining Guo.
\newblock Styleswin: Transformer-based gan for high-resolution image generation.
\newblock In \emph{Proceedings of the IEEE/CVF conference on computer vision and pattern recognition}, pages 11304--11314, 2022.

\bibitem[Wang et~al.(2019)Wang, Sun, Liu, Sarma, Bronstein, and Solomon]{wang2019dynamic}
Yue Wang, Yongbin Sun, Ziwei Liu, Sanjay~E Sarma, Michael~M Bronstein, and Justin~M Solomon.
\newblock Dynamic graph cnn for learning on point clouds.
\newblock \emph{ACM Transactions on Graphics (tog)}, 38\penalty0 (5):\penalty0 1--12, 2019.

\bibitem[Shi et~al.(2016)Shi, Caballero, Husz{\'a}r, Totz, Aitken, Bishop, Rueckert, and Wang]{shi2016real}
Wenzhe Shi, Jose Caballero, Ferenc Husz{\'a}r, Johannes Totz, Andrew~P Aitken, Rob Bishop, Daniel Rueckert, and Zehan Wang.
\newblock Real-time single image and video super-resolution using an efficient sub-pixel convolutional neural network.
\newblock In \emph{Proceedings of the IEEE conference on computer vision and pattern recognition}, pages 1874--1883, 2016.

\bibitem[Touvron et~al.(2021)Touvron, Cord, Sablayrolles, Synnaeve, and J{\'e}gou]{touvron2021going}
Hugo Touvron, Matthieu Cord, Alexandre Sablayrolles, Gabriel Synnaeve, and Herv{\'e} J{\'e}gou.
\newblock Going deeper with image transformers.
\newblock In \emph{Proceedings of the IEEE/CVF international conference on computer vision}, pages 32--42, 2021.

\bibitem[Peebles and Xie(2023)]{peebles2023scalable}
William Peebles and Saining Xie.
\newblock Scalable diffusion models with transformers.
\newblock In \emph{Proceedings of the IEEE/CVF International Conference on Computer Vision}, pages 4195--4205, 2023.

\bibitem[Goodfellow et~al.(2014)Goodfellow, Pouget-Abadie, Mirza, Xu, Warde-Farley, Ozair, Courville, and Bengio]{goodfellow2014generative}
Ian Goodfellow, Jean Pouget-Abadie, Mehdi Mirza, Bing Xu, David Warde-Farley, Sherjil Ozair, Aaron Courville, and Yoshua Bengio.
\newblock Generative adversarial nets.
\newblock \emph{Advances in neural information processing systems}, 27, 2014.

\bibitem[Mescheder et~al.(2018)Mescheder, Geiger, and Nowozin]{mescheder2018training}
Lars Mescheder, Andreas Geiger, and Sebastian Nowozin.
\newblock Which training methods for gans do actually converge?
\newblock In \emph{International conference on machine learning}, pages 3481--3490. PMLR, 2018.

\bibitem[Choi et~al.(2020)Choi, Uh, Yoo, and Ha]{choi2020stargan}
Yunjey Choi, Youngjung Uh, Jaejun Yoo, and Jung-Woo Ha.
\newblock Stargan v2: Diverse image synthesis for multiple domains.
\newblock In \emph{Proceedings of the IEEE/CVF conference on computer vision and pattern recognition}, pages 8188--8197, 2020.

\bibitem[Karras et~al.(2020{\natexlab{b}})Karras, Aittala, Hellsten, Laine, Lehtinen, and Aila]{karras2020training}
Tero Karras, Miika Aittala, Janne Hellsten, Samuli Laine, Jaakko Lehtinen, and Timo Aila.
\newblock Training generative adversarial networks with limited data.
\newblock \emph{Advances in neural information processing systems}, 33:\penalty0 12104--12114, 2020{\natexlab{b}}.

\bibitem[Deng et~al.(2019)Deng, Yang, Xu, Chen, Jia, and Tong]{deng2019accurate}
Yu~Deng, Jiaolong Yang, Sicheng Xu, Dong Chen, Yunde Jia, and Xin Tong.
\newblock Accurate 3d face reconstruction with weakly-supervised learning: From single image to image set.
\newblock In \emph{Proceedings of the IEEE/CVF conference on computer vision and pattern recognition workshops}, pages 0--0, 2019.

\bibitem[Lee(2018)]{catpose}
T.~B. Lee.
\newblock Cat hipsterizer,.
\newblock \url{https://github.com/kairess/cat_hipsterizer}, 2018.

\bibitem[Zhao et~al.(2022)Zhao, Ma, G{\"u}era, Ren, Schwing, and Colburn]{zhao2022generative}
Xiaoming Zhao, Fangchang Ma, David G{\"u}era, Zhile Ren, Alexander~G Schwing, and Alex Colburn.
\newblock Generative multiplane images: Making a 2d gan 3d-aware.
\newblock In \emph{European Conference on Computer Vision}, pages 18--35. Springer, 2022.

\bibitem[Xiang et~al.(2023)Xiang, Yang, Deng, and Tong]{xiang2023gram}
Jianfeng Xiang, Jiaolong Yang, Yu~Deng, and Xin Tong.
\newblock Gram-hd: 3d-consistent image generation at high resolution with generative radiance manifolds.
\newblock In \emph{Proceedings of the IEEE/CVF International Conference on Computer Vision}, pages 2195--2205, 2023.

\bibitem[Yu et~al.(2018)Yu, Wang, Peng, Gao, Yu, and Sang]{yu2018bisenet}
Changqian Yu, Jingbo Wang, Chao Peng, Changxin Gao, Gang Yu, and Nong Sang.
\newblock Bisenet: Bilateral segmentation network for real-time semantic segmentation.
\newblock In \emph{Proceedings of the European conference on computer vision (ECCV)}, pages 325--341, 2018.

\bibitem[Wang et~al.(2023{\natexlab{c}})Wang, Han, Habermann, Daniilidis, Theobalt, and Liu]{wang2023neus2}
Yiming Wang, Qin Han, Marc Habermann, Kostas Daniilidis, Christian Theobalt, and Lingjie Liu.
\newblock Neus2: Fast learning of neural implicit surfaces for multi-view reconstruction.
\newblock In \emph{Proceedings of the IEEE/CVF International Conference on Computer Vision}, pages 3295--3306, 2023{\natexlab{c}}.

\bibitem[Karras et~al.(2017)Karras, Aila, Laine, and Lehtinen]{karras2017progressive}
Tero Karras, Timo Aila, Samuli Laine, and Jaakko Lehtinen.
\newblock Progressive growing of gans for improved quality, stability, and variation.
\newblock \emph{arXiv preprint arXiv:1710.10196}, 2017.

\bibitem[Abdal et~al.(2019)Abdal, Qin, and Wonka]{abdal2019image2stylegan}
Rameen Abdal, Yipeng Qin, and Peter Wonka.
\newblock Image2stylegan: How to embed images into the stylegan latent space?
\newblock In \emph{Proceedings of the IEEE/CVF international conference on computer vision}, pages 4432--4441, 2019.

\bibitem[Roich et~al.(2022)Roich, Mokady, Bermano, and Cohen-Or]{roich2022pivotal}
Daniel Roich, Ron Mokady, Amit~H Bermano, and Daniel Cohen-Or.
\newblock Pivotal tuning for latent-based editing of real images.
\newblock \emph{ACM Transactions on graphics (TOG)}, 42\penalty0 (1):\penalty0 1--13, 2022.

\end{thebibliography}
}

\newpage
\appendix
\section{Appendix}
\label{sec: appendix}

\subsection{Implementation details}
\label{sec: appendix implementation details}
First of all, please refer to the attached code in supplementary for a more detailed implementation of the proposed method.
For the coefficients of objective functions, we use $\lambda=1$, $\lambda_{\text{pose}}=1$, and $\lambda_{\text{knn}}=10$.
We train the model until the discriminator sees 10-15M images.
Differently, we use $\lambda_{\text{center}}=1, 10$ each for FFHQ and AFHQ-Cat datasets.
Following prior works~\cite{chan2022efficient,chen2023mimic3d}, we also generator pose conditioning~\cite{chan2022efficient} for modeling pose-specific attributes.
Other hyperparameters not mentioned are identical to EG3D~\cite{chan2022efficient}.

For the architectural details, we use an upsampling ratio $r=4$ and the initial number of Gaussians $N=256$.
For the number of hierarchical levels, we adopt $L=5$ for the dataset with 256$\times$256 resolution, and $L=6$ for 512$\times$512 resolution.
We use the stack of 6 $\texttt{block}$s at the lowest level and 1 $\texttt{block}$ for the other levels, that is, 87K and 349K Gaussians are used for 256 and 512 resolutions.
The number of channels is [512, 512, 256, 128, 128, 128] for the block of levels [0, 1, 2, 3, 4, 5], respectively.
We use equalized learning rate~\cite{karras2017progressive} for every layer.
Also, we only apply AdaIN and adaptive layerscale on the layers lower than level 3, as we observe the modulation on higher levels does not affect the rendering results much.
For discriminator architecture, we use StyleGAN2 discriminator following EG3D~\cite{chan2022efficient}, but with an additional pose embedding layer composed of the MLP layer.

For the representation of Gaussian parameters, we follow the 3D-GS~\cite{kerbl20233d} that uses the exponential activation function for the scale parameters and normalizes the quaternion to make it valid.
As these operations are performed in the rendering process, we omit them for simplicity.

\paragraph{Attention layer}
We deploy the attention layer for modeling the interaction between the Gaussians.
For the lowest levels $l=0$, we utilize the general self-attention.
Since the self-attention on a large number of Gaussians requires too much computation cost, we use local attention similar to EdgeConv~\cite{wang2019dynamic}, which computes attention scores against KNN points~($k=16$),  for the levels $l > 0$.
As calculating the KNN points also becomes demanding where the number of points to be enlarged, we model the points to interact with the points from the identical parent point, for levels $l > 3$, instead of computing the actual KNN points.

\paragraph{Output layer~($\texttt{toGaus}$)}
$\texttt{toGaus}$ layer is basically a set of linear layers that synthesize the Gaussian parameter.
That is, it consists of 5 linear layers which are parallelly computed to output $\hat{\mu}, \hat{s}, \hat{\alpha}, \hat{q}, \hat{c}$, respectively.
They have an input as the intermediate feature $x^l$, and an additional camera parameter $\theta$ only for the color $\hat{c}$.
In addition, the layer for the relative position $\hat{\mu}$ has a tanh function as the activation to restrict the range of output position, and the layer for the scale difference $\hat{s}$ has a softplus function multiplied by -1 as the activation to make it have negative values.
Thus, $\texttt{toGaus}$ layer at a specific level is defined as follows:
\begin{equation}
    p^l = \texttt{Lin}_{p}(x^l), \quad 
    \hat{\mu}^l = \text{tanh}(\texttt{Lin}_{\hat{\mu}}(x^l)), \quad
    \hat{s}^l = -\text{softplus}(\texttt{Lin}_{\hat{s}}(x^l)), \quad
    \hat{c}^l = \texttt{Lin}_{\hat{c}}(x^l, \theta), 
\end{equation}
where $p \in \{ \hat{\alpha}, \hat{q} \}$ and $\texttt{Lin}_{(\cdot)}$ is a linear layer for the Gaussian parameter.
Also, we initialize the bias of linear layers to set the quaternion to be an identity rotation matrix and the opacity to be 0.1 and the scale to be $e^{-\frac{1}{\sqrt{N}}}$.

The constant $\Delta s$ reducing the scale along with hierarchy is adjusted by the resolution of the given data as below:
\begin{equation}
    \Delta s = - \log (\sqrt{H \times W} / (L \times \sqrt{N})),
\end{equation}
where $H, W$ are the height and width of the image, $L$ is the total level of hierarchy, and $N$ is the initial number of Gaussians at level $l=0$.

\paragraph{Background generator}
The architecture of the background generator is similar to the proposed generator, but the architecture of $\texttt{block}$ is different.
In detail, we do not utilize the attention and MLP layers, instead using simple linear layers with demod- and modulation~\cite{karras2019style} to reduce the computational cost.
Also, the position parameters synthesized by the background generator are normalized to make the generated Gaussians located on the sphere.
We set the radius of the sphere as 3.

For the hyperparameters, we set the number of initial Gaussians $N=2000$, and the number of hierarchy $L=2$ with upsampling factor $r=4$ and the channel of every layer as 128. 
Thus, the background is composed of 10,000 Gaussians in total.

\paragraph{Training trick for FFHQ dataset}
We observe that training on the FFHQ dataset lacks the capability to accurately model the precise structure of 3D scenes, particularly for structures viewed from different vertical angles. To address this issue, we augment the pose distribution during training. Specifically, with a probability of 0.5, we sample poses from a pre-defined distribution characterized by the standard deviation of yaw and pitch~($\sigma_{\text{yaw}}, \sigma_{\text{pitch}}$) of the FFHQ dataset. We define the yaw distribution as $N(0, \sigma_{\text{yaw}})$ and the pitch distribution as $U(-3 \sigma_{\text{pitch}}, 3 \sigma_{\text{pitch}})$, where $U$ represents a uniform distribution. This oversampling strategy ensures better coverage of images with varied vertical perspectives.

\subsection{Additional examples}
\label{sec: appendix additional examples}
\begin{figure}
    \centering
    \includegraphics[width=\textwidth]{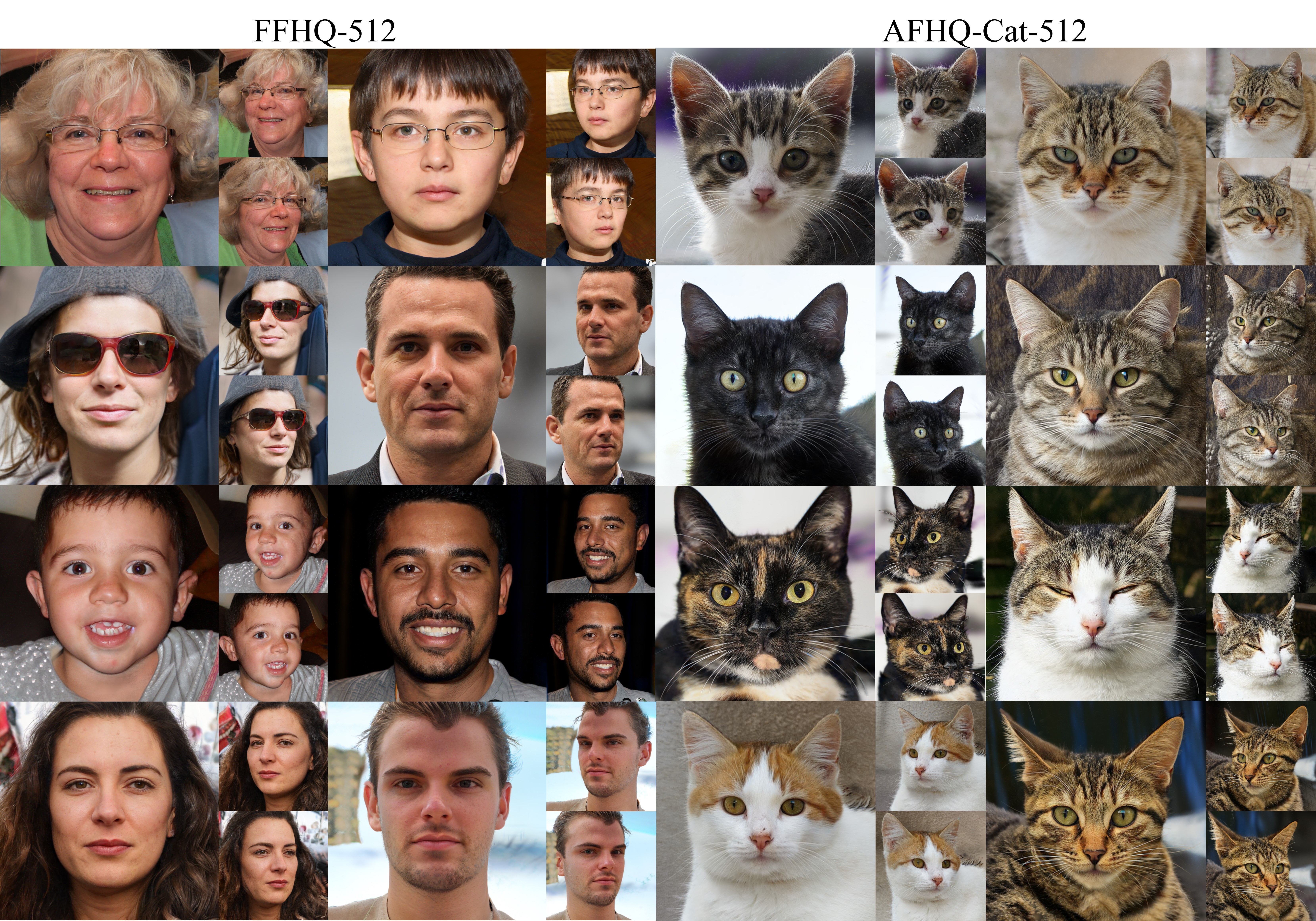}
    \caption{Generated examples on FFHQ-512 and AFHQ-Cat-512}
    \label{fig: appendix_examples_512}
\end{figure}

\begin{figure}
    \centering
    \includegraphics[width=\textwidth]{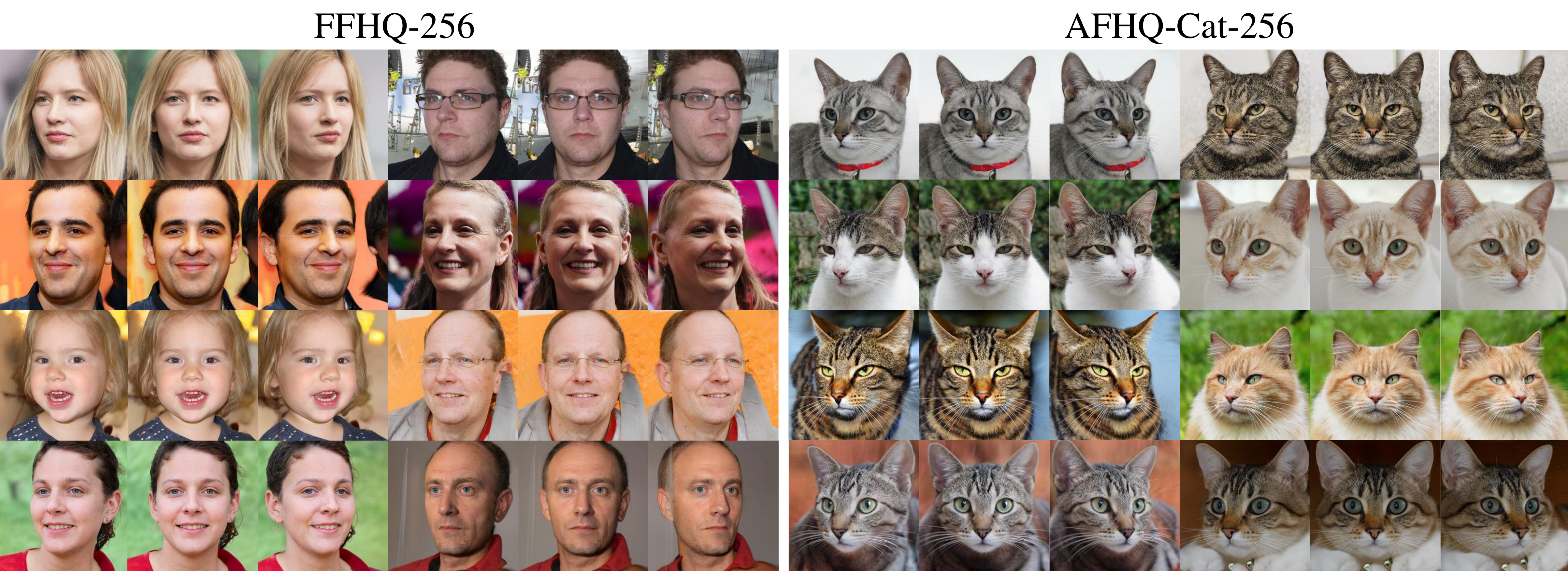}
    \caption{Generated examples on FFHQ-256 and AFHQ-Cat-256}
    \label{fig: appendix_examples_256}
\end{figure}

\begin{figure}
    \centering
    \includegraphics[width=1.0\textwidth]{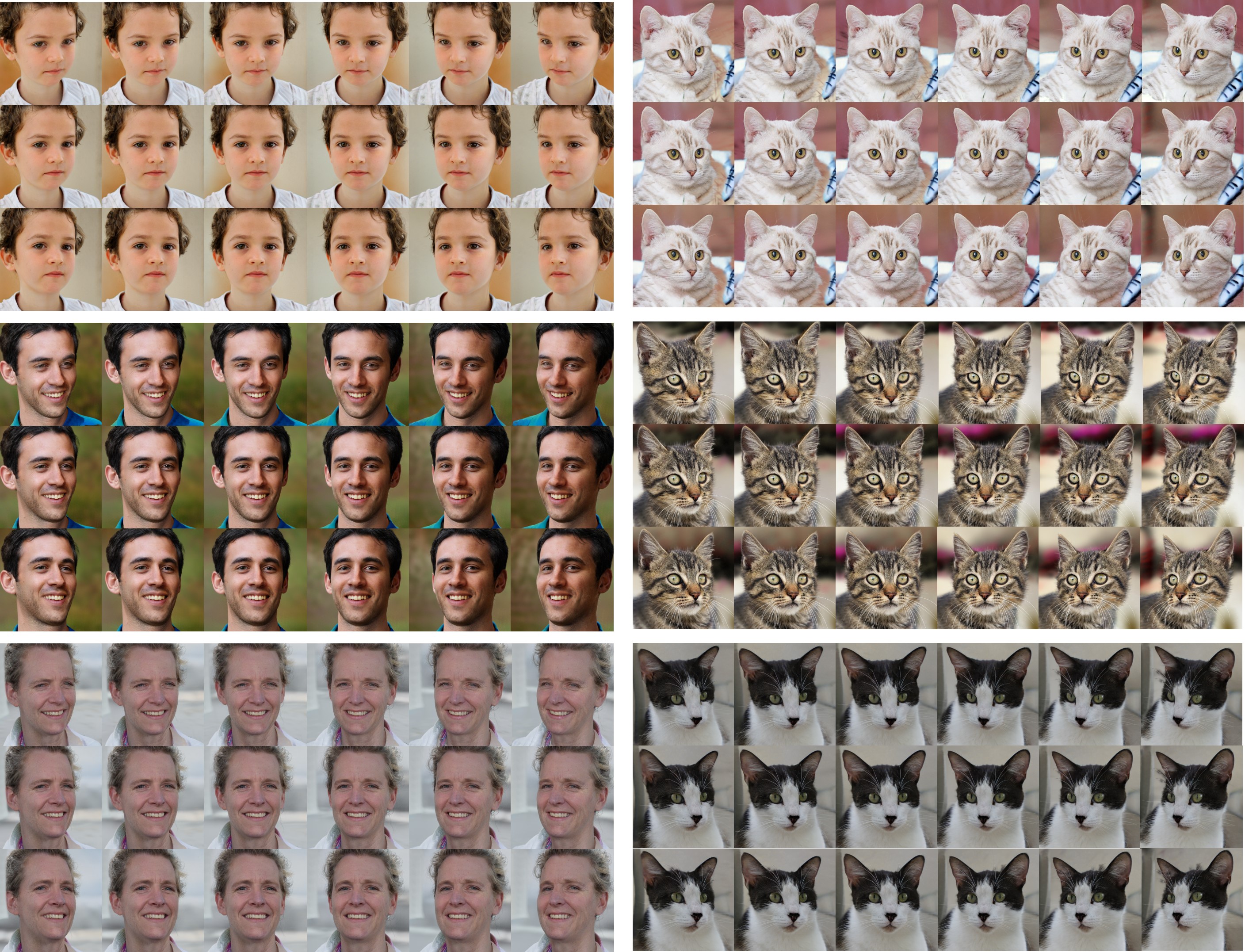}
    \caption{Examples of multi-view generation. We sample the camera pose of [-0.4, 0.4] radian for a yaw, and [-0.4, +0.1] for a pitch.}
    \label{fig: appendix_multiview_dense}
\end{figure}

We present additional examples generated by our method for the FFHQ and AFHQ-Cat datasets in Fig.~\ref{fig: appendix_examples_512},~\ref{fig: appendix_examples_256}.
Furthermore, we provide multi-view images synthesized from densely sampled camera positions in Fig.~\ref{fig: appendix_multiview_dense}. Specifically, for each 3D scene, we sample yaw angles in the range of [-0.4, 0.4] radians and pitch angles in the range of [-0.4, 0.1] radians. These multi-view images demonstrate the 3D consistency of our method in a qualitative manner.
For every visualization, we use truncation psi $\psi=0.7$ and fix a camera condition as frontal view.

\subsection{Latent interpolation}

\begin{figure}
    \centering
    \includegraphics[width=\textwidth]{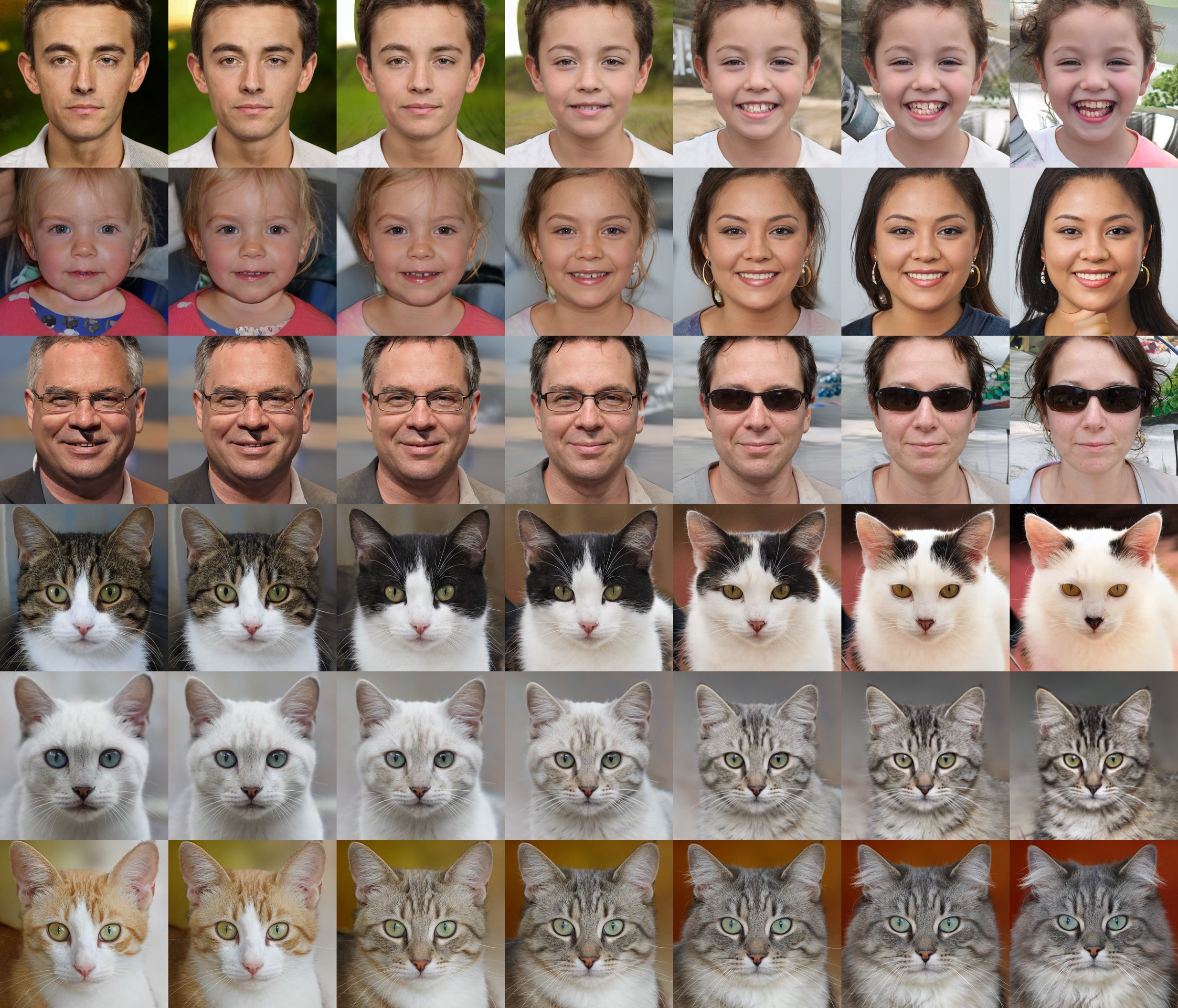}
    \caption{Linear interpolation between latent codes~($w$ space)}
    \label{fig: appendix_winterp}
\end{figure}
One of the notable characteristics of GANs is a semantic latent space. To validate that the proposed method also exhibits this characteristic, we synthesize images using linearly interpolated latent codes. As shown in Fig.~\ref{fig: appendix_winterp}, the interpolated images gradually transition from the given source to the target image, demonstrating smooth and coherent transformations. This indicates that our method maintains a meaningful and semantically rich latent space.

\subsection{w+ Inversion}
\begin{figure}
    \centering
    \includegraphics[width=0.9\textwidth]{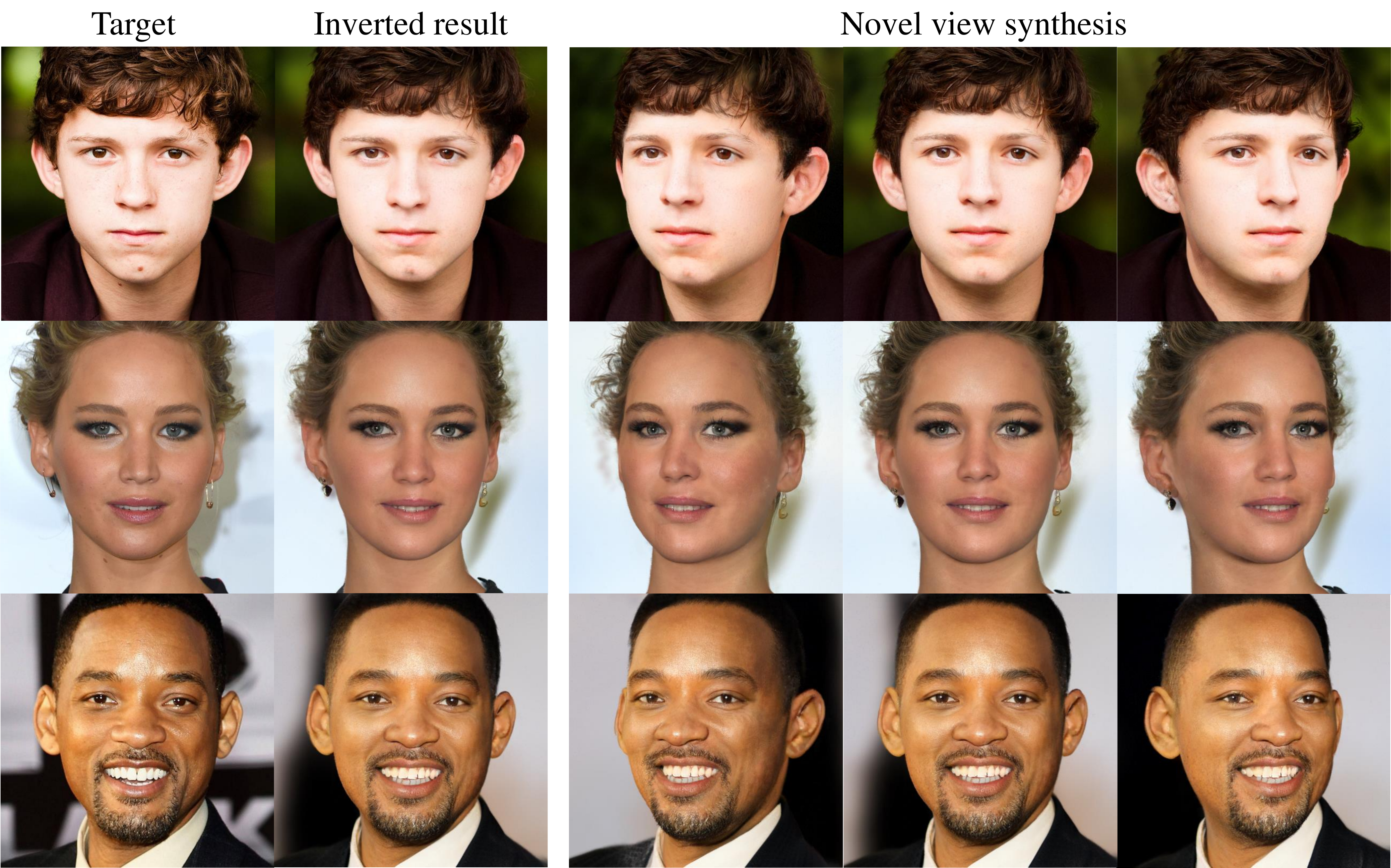}
    \caption{Novel view synthesis using w+ inversion.}
    \label{fig: appendix inversion}
\end{figure}

We perform the w+ inversion~\cite{abdal2019image2stylegan} from a given single view in-the-wild image and predict novel view synthesis using the estimated 3D scene. As shown in Fig.~\ref{fig: appendix inversion}, the proposed method successfully inverts the in-the-wild image and synthesizes novel views from it, demonstrating the method's ability to generalize and manipulate real-world images. We expect that recent inversion techniques, such as pivotal tuning~\cite{roich2022pivotal}, could further enhance the quality of the inversion.

\subsection{Visualization of anchor Gaussians}

\begin{figure}
    \centering
    \includegraphics[width=0.9\textwidth]{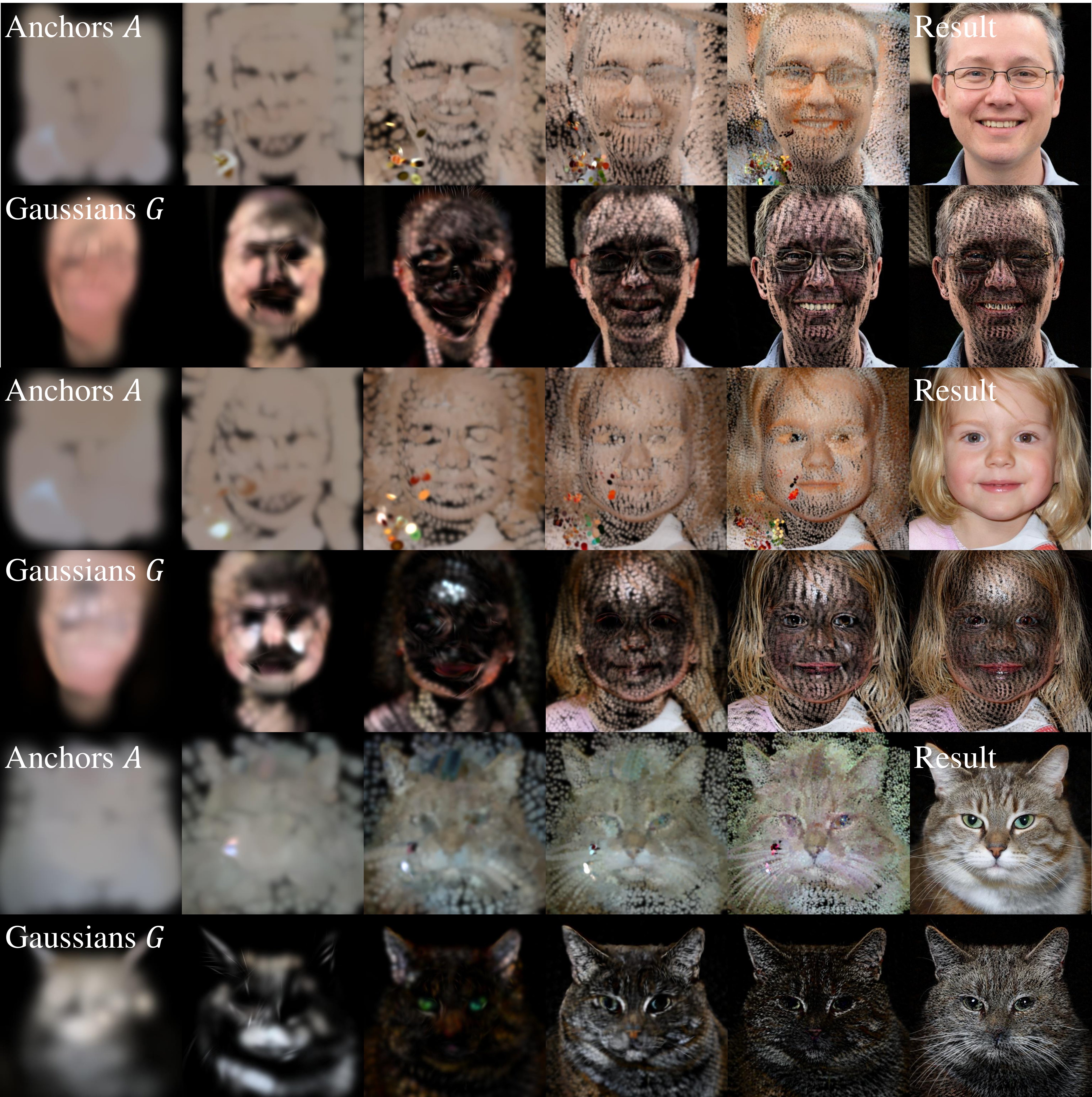}
    \caption{Visualization of Gaussians $G$ and anchor Gaussians $A$ for each level. We set opacity $\alpha$ of anchors as $\text{sigmoid}(1)$ for visualization.
    From left-to-right, levels of Gaussians and anchors increase except right-top image, which is a composite of Gaussians from all levels.
    }
    \label{fig: appendix_anchors}
\end{figure}
We visualize the Gaussians and anchor Gaussians level-by-level in Fig.~\ref{fig: appendix_anchors}.
Note that, we manually set the opacity of the anchors as $\text{sigmoid}(1)$, as they do not have actual opacity for rendering.
As visualized, we observe that the anchors typically have a larger scale compared to the Gaussians.
We also verify it by calculating statistics of the scale of anchors and Gaussians.
In detail, the average scale of anchors is $e^{-5.08}$ across 100 randomly sampled scenes, while the average scale of Gaussians is $e^{-7.12}$.
We hypothesize that the Gaussians adjust their scales to represent sharp details in real-world images, whereas the anchor Gaussians are mainly used to constrain and regularize the parameters of the coarser-level Gaussians.

\subsection{Effect of background generator}

\begin{figure}
    \centering
    \includegraphics[width=0.7\textwidth]{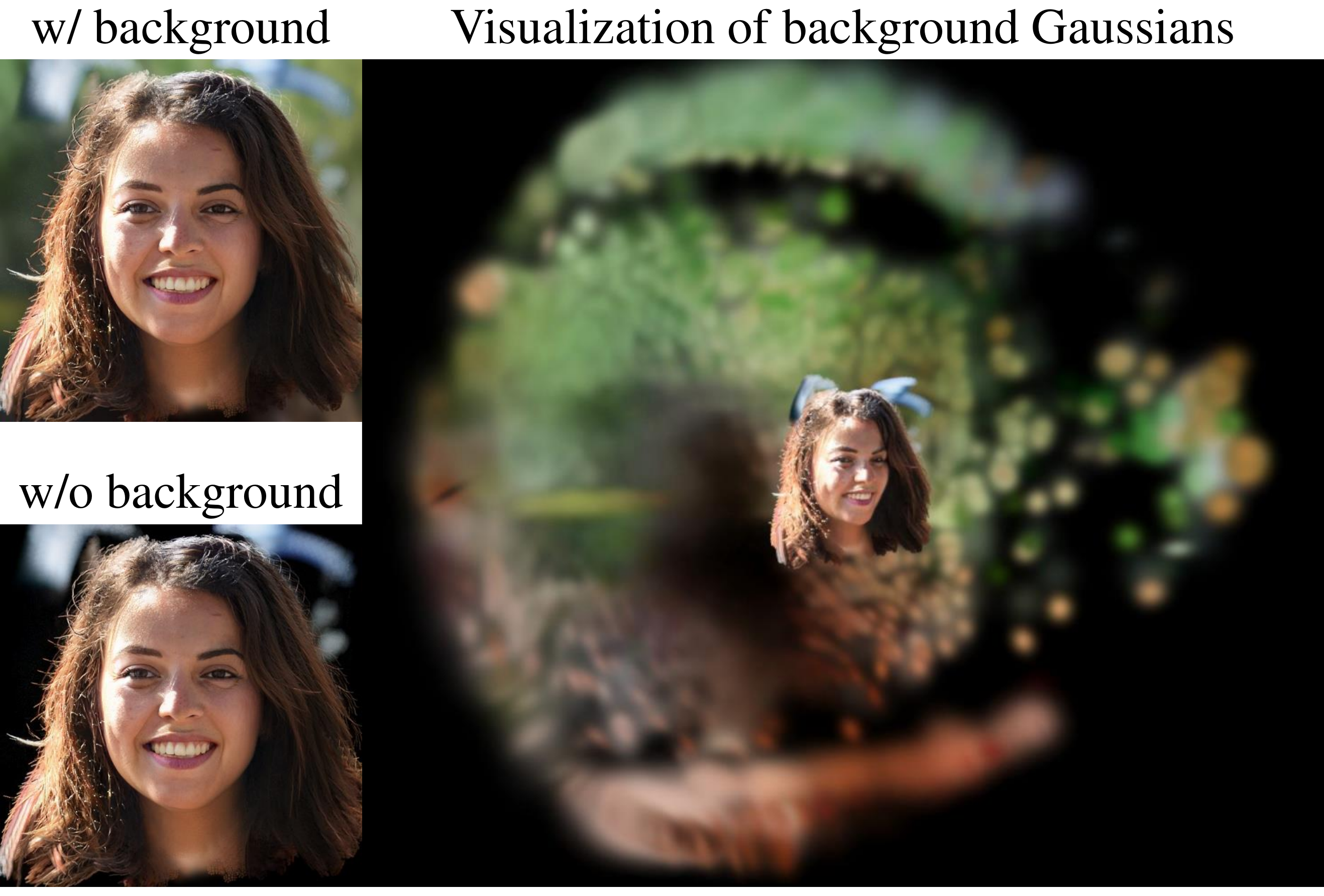}
    \caption{Visualization of effects of background generator}
    \label{fig: appendix_background_gen}
\end{figure}
We visualize the Gaussians generated from the background generator in Fig.~\ref{fig: appendix_background_gen}.
On the left side, we synthesize the image with and without the background Gaussians.
As visualized, the background Gaussians are responsible for the backgrounds, especially for the blurred area.
For more visualization, we provide an image rendered by a camera far from the world-coordinate origin on the right side.

\clearpage


\newpage
\section*{NeurIPS Paper Checklist}

\begin{enumerate}

\item {\bf Claims}
    \item[] Question: Do the main claims made in the abstract and introduction accurately reflect the paper's contributions and scope?
    \item[] Answer: \answerYes{}
    \item[] Justification: We elaborate on our claims in both abstract and introduction.

\item {\bf Limitations}
    \item[] Question: Does the paper discuss the limitations of the work performed by the authors?
    \item[] Answer: \answerYes{}
    \item[] Justification: We refer to the limitation on sec.~\ref{sec: limitation}.

\item {\bf Theory Assumptions and Proofs}
    \item[] Question: For each theoretical result, does the paper provide the full set of assumptions and a complete (and correct) proof?
    \item[] Answer: \answerNA{}
    \item[] Justification: This paper does not contain any theory assumptions.

\item {\bf Experimental Result Reproducibility}
    \item[] Question: Does the paper fully disclose all the information needed to reproduce the main experimental results of the paper to the extent that it affects the main claims and/or conclusions of the paper (regardless of whether the code and data are provided or not)?
    \item[] Answer: \answerYes{}
    \item[] Justification: We report the implementation details in sec.~\ref{sec: appendix implementation details} and reference of datasets in sec.~\ref{sec:4.1 experimental settings}, and also attach the implementation code in supplementary.

\item {\bf Open access to data and code}
    \item[] Question: Does the paper provide open access to the data and code, with sufficient instructions to faithfully reproduce the main experimental results, as described in supplemental material?
    \item[] Answer: \answerYes{}
    \item[] Justification: We attach the implementation code in supplementary and will release it.

\item {\bf Experimental Setting/Details}
    \item[] Question: Does the paper specify all the training and test details (e.g., data splits, hyperparameters, how they were chosen, type of optimizer, etc.) necessary to understand the results?
    \item[] Answer: \answerYes{}
    \item[] Justification: We report the detailed training and test details in appendix~\ref{sec: appendix}.

\item {\bf Experiment Statistical Significance}
    \item[] Question: Does the paper report error bars suitably and correctly defined or other appropriate information about the statistical significance of the experiments?
    \item[] Answer: \answerNo{} 
    \item[] Justification: Multiple runs of training are computationally too demanding in 3D GANs.

\item {\bf Experiments Compute Resources}
    \item[] Question: For each experiment, does the paper provide sufficient information on the computer resources (type of compute workers, memory, time of execution) needed to reproduce the experiments?
    \item[] Answer: \answerYes{}
    \item[] Justification: We report the training resource in Sec.~\ref{sec:4 experiments}, which is in charge of most of the computational cost for the experiments in this paper.
    
\item {\bf Code Of Ethics}
    \item[] Question: Does the research conducted in the paper conform, in every respect, with the NeurIPS Code of Ethics \url{https://neurips.cc/public/EthicsGuidelines}?
    \item[] Answer: \answerYes{}
    \item[] Justification: We review the NeurIPS Code of Ethics.

\item {\bf Broader Impacts}
    \item[] Question: Does the paper discuss both potential positive societal impacts and negative societal impacts of the work performed?
    \item[] Answer: \answerYes{}
    \item[] Justification: We elaborate on the broader impacts in sec.~\ref{sec: broader impact}.
    
\item {\bf Safeguards}
    \item[] Question: Does the paper describe safeguards that have been put in place for responsible release of data or models that have a high risk for misuse (e.g., pretrained language models, image generators, or scraped datasets)?
    \item[] Answer: \answerNA{}
    \item[] Justification: The paper does not have such risks, as we do not use models and data concerned about any privacy issues.

\item {\bf Licenses for existing assets}
    \item[] Question: Are the creators or original owners of assets (e.g., code, data, models), used in the paper, properly credited and are the license and terms of use explicitly mentioned and properly respected?
    \item[] Answer: \answerYes{}
    \item[] Justification: We check datasets has license of CC BY-NC-SA 4.0 and CC-BY-NC 4.0.

\item {\bf New Assets}
    \item[] Question: Are new assets introduced in the paper well documented and is the documentation provided alongside the assets?
    \item[] Answer: \answerNA{}
    \item[] Justification: We do not provide new assets.

\item {\bf Crowdsourcing and Research with Human Subjects}
    \item[] Question: For crowdsourcing experiments and research with human subjects, does the paper include the full text of instructions given to participants and screenshots, if applicable, as well as details about compensation (if any)? 
    \item[] Answer: \answerNA{}
    \item[] Justification: We do not involve crowdsourcing nor research with human subjects.

\item {\bf Institutional Review Board (IRB) Approvals or Equivalent for Research with Human Subjects}
    \item[] Question: Does the paper describe potential risks incurred by study participants, whether such risks were disclosed to the subjects, and whether Institutional Review Board (IRB) approvals (or an equivalent approval/review based on the requirements of your country or institution) were obtained?
    \item[] Answer: \answerNA{}
    \item[] Justification: We do not involve crowdsourcing nor research with human subjects.

\end{enumerate}

\end{document}